\documentclass{article}

\usepackage{microtype}
\usepackage{graphicx}
\usepackage{subfigure}
\usepackage{booktabs} 

\usepackage{hyperref}

\usepackage[accepted]{icml2025}

\usepackage{amsmath}
\usepackage{amssymb}
\usepackage{mathtools}
\usepackage{amsthm}

\usepackage{wrapfig}
\usepackage{caption}
\usepackage{listings}
\usepackage{adjustbox}
\usepackage[capitalize,noabbrev]{cleveref}

\theoremstyle{plain}

\theoremstyle{definition}

\theoremstyle{remark}

\usepackage[textsize=tiny]{todonotes}
\usepackage{placeins}
\usepackage{siunitx}

\icmltitlerunning{Beyond Fine-Tuning: A Systematic Study of Sampling Techniques in Personalized Image Generation}

\begin{document}

\twocolumn[
\icmltitle{Beyond Fine-Tuning: \\ A Systematic Study of Sampling Techniques in \\ Personalized Image Generation}



\icmlsetsymbol{equal}{*}

\begin{icmlauthorlist}
\icmlauthor{Vera Soboleva}{equal,yyy,comp}
\icmlauthor{Maksim Nakhodnov}{equal,comp}
\icmlauthor{Aibek Alanov}{yyy,comp}
\end{icmlauthorlist}

\icmlaffiliation{yyy}{HSE University, Moscow, Russia}
\icmlaffiliation{comp}{AIRI, Moscow, Russia}

\icmlcorrespondingauthor{Aibek Alanov}{alanov.aibek@gmail.com}

\icmlkeywords{Machine Learning, ICML}

\vskip 0.3in
]



\printAffiliationsAndNotice{\icmlEqualContribution} 


\begin{abstract}
Personalized text-to-image generation aims to create images tailored to user-defined concepts and textual descriptions. Balancing the fidelity of the learned concept with its ability for generation in various contexts presents a significant challenge. Existing methods often address this through diverse fine-tuning parameterizations and improved sampling strategies that integrate superclass trajectories during the diffusion process. While improved sampling offers a cost-effective, training-free solution for enhancing fine-tuned models, systematic analyses of these methods remain limited. Current approaches typically tie sampling strategies with fixed fine-tuning configurations, making it difficult to isolate their impact on generation outcomes. To address this issue, we systematically analyze sampling strategies beyond fine-tuning, exploring the impact of concept and superclass trajectories on the results. Building on this analysis, we propose a decision framework evaluating text alignment, computational constraints, and fidelity objectives to guide strategy selection. It integrates with diverse architectures and training approaches, systematically optimizing concept preservation, prompt adherence, and resource efficiency. The source code can be found
at \href{https://github.com/ControlGenAI/PersonGenSampler}{github.com/ControlGenAI/PersonGenSampler}.
\end{abstract}

\section{Introduction}
\label{sec:intro}


Diffusion-based text-to-image generation models~\citep{ramesh2022hierarchical, saharia2022photorealistic, rombach2022high}, trained on large datasets, have recently achieved impressive results in generating photorealistic images from textual prompts. Despite their advanced performance, these models have limitations in generating user-defined concepts, that are difficult to describe accurately using text alone. This limitation has led to increased interest in subject-driven text-to-image generation \citep{DB, TI}. In this task, given a small image dataset (3-5 images) of a given subject, we want to introduce the knowledge of this subject into the pre-trained text-to-image diffusion model and learn to generate it in different contexts described by textual prompts.


Balancing the preservation of a concept's identity with its adaptation to a new context is the main challenge in personalized image generation. On the one hand, the model must generate high-fidelity images of the concepts, even if it has never encountered them during the pre-training phase. On the other hand, the model should not overfit to retain the ability to follow different textual descriptions of the scenes.

Modern techniques introduce various improvements to the training process to balance concept fidelity and editability. These include fine-tuning parameterizations \citep{DB, TI, CD, svdiff, r1e, ortogonal}, regularizations \citep{DB, CD}, and encoder-based paradigms \citep{elite}. For a more comprehensive overview, see Appendix~\ref{app:related_work}. Another direction is to use sampling methods applied after training to improve an already fine-tuned model. The main idea of such methods~\citep{profusion, photoswap} is to combine the sampling trajectories of prompts with concept and superclass tokens (e.g., for a dog concept, we mix trajectories for two prompts: \textit{"a purple V*"} and \textit{"a purple dog"}, see Figure~\ref{fig:sampling}). Sampling-based approaches can provide a cost-effective, training-free way to improve the balance between concept identity and editability. Although fine-tuning and sampling are two distinct strategies to address the same issue, current research frequently overlooks the differences between these methodologies. For example, current works~\citep{profusion, photoswap} introduce complex sampling procedures alongside fixed fine-tuning, leaving unclear the impact of sampling on generation, as does its compatibility with other fine-tuning strategies. Furthermore, they do not compare the proposed strategies with naive sampling approaches, resulting in a lack of understanding of how superclass trajectories influence the sampling process.
In summary, the personalized generation sampling process remains underexplored, with three main open challenges: (1) \textit{The impact of superclass trajectory integration is under-researched}, as previous work has not fully elucidated how the incorporation of superclass trajectories affects the generation output. (2) \textit{Limitations imposed by fine-tuning strategies}; current sampling methods are almost always tied to specific fine-tuning schemes, which limits the ability to study sampling independently and hampers fair comparisons between different approaches. (3) \textit{Simple sampling baselines are often overlooked}, and their potential remains undervalued.


To address these challenges, we propose several contributions to advance the understanding and application of sampling strategies in personalized text-to-image generation. Our work explores the impact of sampling methods beyond fine-tuning, establishing simple yet powerful baselines. In particular, we make the following key contributions:

\textbf{1. A systematic and comprehensive analysis of how superclass trajectories influence the sampling process.} We investigate various combinations of concept and superclass trajectories, including Switching, Mixed, and Masked sampling techniques, along with their hybrid variants. We carefully ablate hyperparameters across all methods, assess their importance, and retain only the most impactful ones.


\textbf{2. A finetuning-independent evaluation of various sampling strategies.} We compare various sampling methods, including naive approaches, applied to a fixed fine-tuned model, to analyze the impact of the sampling beyond the fine-tuning strategy. Moreover, we demonstrate how to effectively apply these strategies across architectures and different fine-tuning methods, including various parameterizations, text embedding optimization, and hypernetworks.


\textbf{3. A framework for selecting the most appropriate sampling method for specific generation tasks.} We conduct a fair comparison of sampling methods based on trade-offs between concept fidelity, adaptability, and computational efficiency, and build a framework for identifying the most appropriate sampling method for specific scenarios.

\section{Preliminaries}
\textbf{Stable Diffusion Model} As a base model in this work, we use Stable Diffusion~\citep{stablediffusion}, one of the most widely used diffusion models in research. Stable Diffusion is a large text-to-image model trained on pairs $(x, P)$, where $x$ is an image and $P$ is a text prompt describing it. Stable Diffusion includes the CLIP~\citep{clip} text encoder $E_T$, which is used to obtain the text conditional embedding $p = E_T(P)$, the encoder $E$, which transforms the input image into the latent space $z = E(x)$, the decoder $D$, which reconstructs the input image from the latent $x \approx D(z)$, and a U-Net-based~\citep{unet} conditional diffusion model $\varepsilon_{\theta}$. The denoising process is performed in the latent space. With a randomly sampled noise $\varepsilon \sim N(0,I)$, the time step $t$, and the coefficients controlling the noise schedule we obtain a noisy latent code: $z_t = \alpha_t z + \sigma_t \varepsilon$. The goal of U-Net $\varepsilon_{\theta}$ is to predict the noise from the noisy latent. We assume that $\varepsilon_{\theta}$ depends implicitly on $z_{t}$, although we will omit the dependency in our notation.
\begin{equation}\label{eq:training}
\min_\theta \mathbb{E}_{p, z, \varepsilon, t}\left[\left\|\varepsilon-\varepsilon_\theta(p)\right\|_2^2\right]
\end{equation}
During inference, a random noise $z_T \sim N(0,I)$ is denoised step by step to $z_0$, using DDIM sampling~\cite{ddim}:
$z_{t-1}=\text{DDIM}(t, z_t, \varepsilon_\theta(p)), \; t = T, \dots, 1$. The resulting image is obtained through the decoder as $D(z_0)$. 

\textbf{Classifier-free guidance} A commonly used technique to improve generation quality of conditional diffusion models post-training is classifier-free sampling~\citep{classfree}. Given the current noisy sample $z_t$ and condition $p$, the diffusion model outputs the predictions of the conditional noise $\varepsilon_{\theta}(E_T(p))$ and unconditional noise $\varepsilon_{\theta}$ (conditioned on null text). Then, an updated prediction
\begin{equation}\label{eq:sampling}
\tilde{\varepsilon}_{\theta}(p) = \varepsilon_{\theta} + \omega( \varepsilon_{\theta}(p) - \varepsilon_{\theta}) = \varepsilon_{\theta} + \omega \Delta \varepsilon_{\theta}^{p}
\end{equation}
will be used to sample $z_{t-1}$, where $\omega$ is a guidance scale. $\Delta\varepsilon_{\theta}^{p}$ denotes the classifier-free guidance noise difference.

\textbf{Finetuning for Personalized Text-to-Image Generation} Let $\mathbb{C} = \{ x\}_{i=1}^N$ be a small image set of images with a specific concept. A special text token $V^*$ can be bound to it, using the following fine-tuning objective:
\begin{equation}\label{eq:finetuning}
\min_\theta \mathbb{E}_{z=\mathcal{E}(x), x \in \mathbb{C}, \varepsilon, t}\left[\left\|\varepsilon-\varepsilon_\theta(p^{C})\right\|_2^2\right]
\end{equation}
where $p^C = E_T(P^C)$ is a text embedding of the prompt $P^C = $\textit{"a photo of a V*"}.
\section{Methods}
\label{sec:method}
Given a model $\varepsilon_{\theta}$, fine-tuned by (\ref{eq:finetuning}) for a specific concept, we can identify two distinct sampling approaches, each maximizing one of the objectives: concept fidelity or editability:

Sampling with concept (Base sampling):
\begin{equation} 
  \label{eq:concept_sampling}
  \tilde{\varepsilon}_{\theta}(p^C) = \varepsilon_{\theta} + \omega(\varepsilon_{\theta}(p^C) - \varepsilon_{\theta}) = \varepsilon_{\theta} + \omega \Delta\varepsilon_{\theta}^{C}
\end{equation}
Sampling with superclass:
\begin{equation}
  \label{eq:superclass_sampling}
  \tilde{\varepsilon}_{\theta}(p^S) = \varepsilon_{\theta} + \omega( \varepsilon_{\theta}(p^S) - \varepsilon_{\theta}) = \varepsilon_{\theta} + \omega \Delta\varepsilon_{\theta}^{S}
\end{equation} 
Here, $p^C$ represents a concept prompt embedding (for example, \textit{"a V* with a city in the background"}), and $p^S$ indicates a superclass prompt embedding (\textit{"a backpack with a city in the background"}) where the concept token $V^*$ has been replaced by a superclass token (\textit{"backpack"}).

\begin{figure*}[ht!]
  \centering
  \vspace{-0.02in}
  \includegraphics[trim={0 7cm 0 7cm},clip,width=\linewidth]{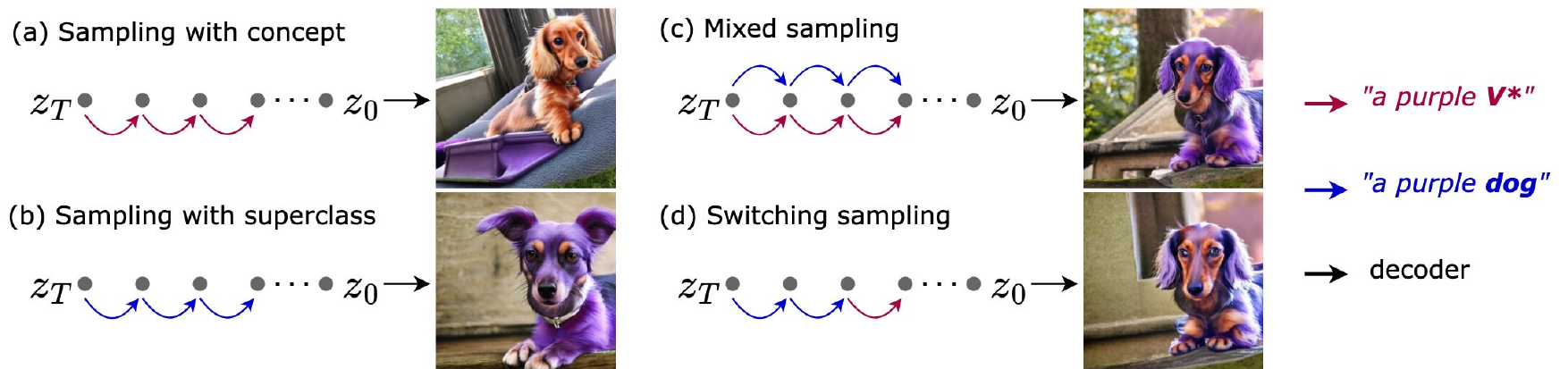}
  \vspace{-0.20in}
  \caption{\textbf{Visualization of Different Sampling Strategies.} (a) Usual sampling with concept reproduces the concept but does not align closely with the text prompt. (b) Generation with superclass effectively captures the context obtained from the prompt but produces a random superclass representative (e.g., dog). (c-d) Mixed and Switching sampling strategies improve context preservation while maintaining the concept's identity.}
  \label{fig:sampling}
  \vspace{-0.20in}
\end{figure*}

The extended fine-tuning of the model \(\varepsilon_{\theta}\) enhances its ability to accurately reproduce the concept generated via~(\ref{eq:concept_sampling}). However, this improvement comes at the cost of overlooking the contextual information supplied by the prompt $P^C$ (see Figure~\ref{fig:sampling}a). Conversely, the generation via~(\ref{eq:superclass_sampling}) ensures the highest alignment with the text prompt, though at the expense of preserving the concept's identity (see Figure~\ref{fig:sampling}b).


This consideration raises the question of whether we can integrate the two sampling strategies~(\ref{eq:concept_sampling}) and~(\ref{eq:superclass_sampling}) to obtain the optimal balance between the high fidelity of the learned concept identity and its adaptability to various contexts.

\subsection{Mixed sampling} \label{sec:mixed_sampling}
One reasonable approach for incorporating superclass into the generation process~\citep{profusion} is to modify the sampling strategy by adding guidance to the superclass prompt (see Figure~\ref{fig:sampling}c):
\begin{align}\label{eq:mixed_sampling}
    \tilde{\varepsilon}^{MX}_{\theta}(p^S, p^C) = \varepsilon_{\theta} + \omega_s \Delta\varepsilon_{\theta}^{S} + \omega_c\Delta\varepsilon_{\theta}^{C}
\end{align}

Adjusting the ratio between the concept guidance scale $\omega_c$ and the superclass guidance scale $\omega_s$ amplifies or diminishes the influence of the concept or superclass, varying the trade-off between concept and context fidelity. Figure~\ref{fig:visual} shows how the generated output changes with increasing superclass influence. For instance, in the teapot example, as we raise the superclass guidance scale, the context, which was initially poorly represented through sampling with the concept, gradually becomes more accurate. However, excessive superclass influence may reduce concept identity preservation, as shown in the dog example.

\subsection{Switching sampling} \label{sec:switching_sampling}
Another solution for how to combine the superclass sampling trajectory with the concept sampling trajectory is to condition several steps on the superclass prompt embedding $p^S$, then at the \textit{switching step} $t_{sw}$ switch to the concept prompt embedding $p^C$  (see Figure~\ref{fig:sampling}d). In this case~(\ref{eq:sampling}) will be rewritten in the following form:
\begin{align}\label{eq:switching_sampling}
    \tilde{\varepsilon}^{SW}_{\theta}(p^S, p^C, t_{sw}) = \varepsilon_{\theta} +
    \begin{cases}
         \omega\Delta\varepsilon_{\theta}^{S}, & t > T - t_{sw}\\
        \omega\Delta\varepsilon_{\theta}^{C}, &\text{otherwise}
    \end{cases} 
\end{align}
By increasing the \textit{switching step} $t_{sw}$, we can amplify the influence of the superclass and thus improve context preservation.
Up to 10 steps can effectively recover context that has been poorly generated through Base sampling, as demonstrated in the teapot example in Figure~\ref{fig:visual}. Nonetheless, this strategy may result in notable degradation of the concept's identity. The effect of the superclass can be so intense that the concept loses its original attributes and takes on excessive characteristics from the superclass, as evidenced by the dog example in Figure~\ref{fig:visual}.


This sampling procedure is similar to Photoswap~\cite{photoswap} but adapted to the personalization task. The main difference is that switched sampling takes noise predictions entirely from the superclass trajectory for the first $t_{sw}$ steps, whereas Photoswap uses only self- and cross-attention maps, and features are taken from the superclass for the first $t_{sw}$ steps. However, as we show in Section~\ref{sec:experiments}, the results of both methods are almost indistinguishable.

The aforementioned methods can be flexibly combined, we refer to this type of sampling as \textit{multi-stage sampling}:
\begin{align}\label{eq:multistage_sampling}
\tilde{\varepsilon}^{MS}_{\theta}(p^S, p^C) = \varepsilon_{\theta} +
    \begin{cases}
         (\omega_s + \omega_c)\Delta\varepsilon_{\theta}^{S} &t > T - t_{sw}\\[-0pt]
         \omega_s \Delta\varepsilon_{\theta}^{S} + \omega_c\Delta\varepsilon_{\theta}^{C}&\text{otherwise}
    \end{cases}
\end{align}
This combination enables a greater influence of the superclass on the generated output and enhances alignment with the text prompt. However, it is important to consider that as the influence of the superclass increases, the more the concept's identity is lost.

\subsection{Masked sampling}

\begin{figure*}[t!]
  \centering
  \includegraphics[trim={0 6.7cm 0 6.7cm},clip,width=\linewidth]{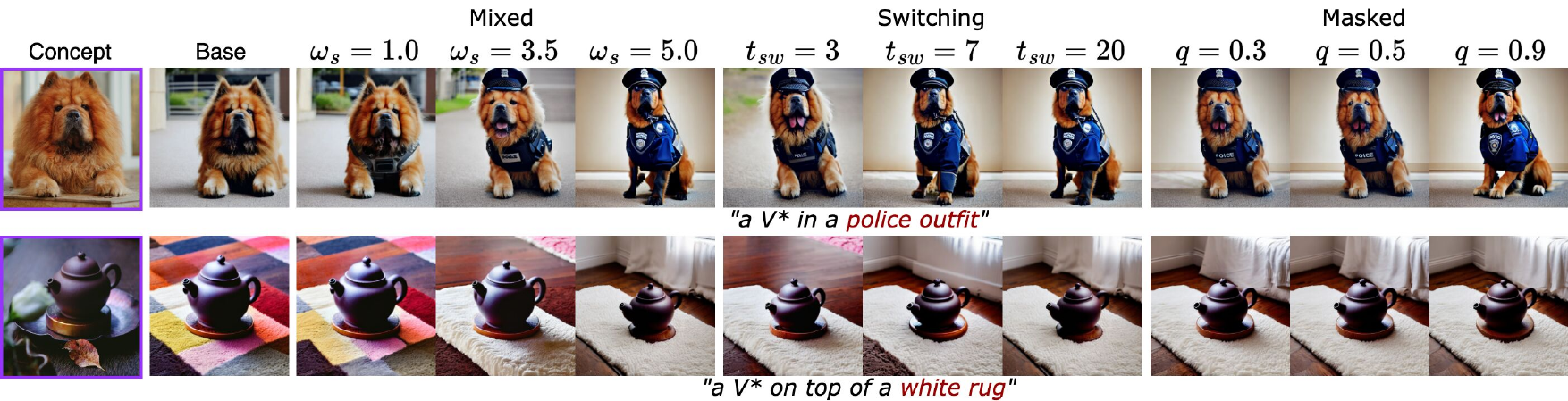}
  \vspace{-0.19in}
  \caption{\textbf{Effects of Superclass Influence on Different Sampling Methods.} 
  For Mixed Sampling, the influence is adjusted by varying the superclass guidance scale $\omega_s$ with $\omega_c = 7.0 - \omega_s$. For Switching Sampling, we vary the switching step $t_{sw}$ . For Masked Sampling, the mask is modified by altering the concept mask thresholding quantile $q$.
  }
  \label{fig:visual}
  \vspace{-0.19in}
\end{figure*}

Sampling with a superclass prompt hinders the preservation of concept identity, whereas sampling with a concept prompt disrupts contextual adaptation. To address this challenge, restricting the image regions impacted by each sampling approach could be beneficial. This can be effectively achieved through masking.

Suppose at each diffusion step we could obtain a concept mask $M_t$, then we can use it in the Mixed sampling. Specifically, we apply this mask to the concept trajectory, ensuring it only influences relevant regions:
\begin{align}\label{eq:masked_base}
    \varepsilon^{M}_{\theta}(p^S, p^C) = \varepsilon_{\theta} + \omega\Delta\varepsilon_{\theta}^{C} \odot M_t + \omega\Delta\varepsilon_{\theta}^{S} \odot \overline{M_t} 
\end{align}
Moreover, to enhance the alignment between regions inside and outside the mask, and to gently amplify the influence of the superclass within the mask -- especially in cases where prompts alter the object's appearance (like color or outfit) -- we can apply Mixed sampling within the mask:
\begin{align}\label{eq:masked_mixed}
    &\varepsilon^{M}_{\theta}(p^S, p^C) = \varepsilon_{\theta} + \\ &+ \omega_c\Delta\varepsilon_{\theta}^{C} \odot M_t + 
    \omega_s\Delta\varepsilon_{\theta}^{S} \odot M_t +(\omega_c + \omega_s)\Delta\varepsilon_{\theta}^{S} \odot \overline{M_t}  \notag
\end{align}
The generation process starts with Mixed sampling for a limited number of steps, thereby enhancing the robustness of mask generation. Then, we apply masked sampling as described in (\ref{eq:masked_mixed}), using the concept mask $M_t(q)$. This mask is derived by averaging the cross-attention maps associated with the concept identifier token across all U-Net layers and binarizing it using a threshold determined by the quantile $q$:
\begin{equation}\label{eq:masked_sampling}
\tilde{\varepsilon}^{M}_{\theta}(p^S, p^C) = 
    \begin{cases}
         \tilde{\varepsilon}^{MX}_{\theta}(p^S, p^C, \omega_c^0, \omega_s^0), & t > T - t_{sw}\\
         \varepsilon^{M}_{\theta}(p^S, p^C, \omega_c, \omega_s, q),
    &\text{otherwise,}
    \end{cases} 
\end{equation}
where $\varepsilon^{M}_{\theta}(p^S, p^C, \omega_c, \omega_s, q)$ is computed as in (\ref{eq:masked_mixed}). 

Equation~\ref{eq:masked_sampling} summarizes the complete Masked sampling algorithm. Increasing the quantile $q$ reduces the area influenced by the concept, thereby expanding the region impacted by the superclass (see Appendix~\ref{app:cross_attn}) and enhancing the influence of the context, as illustrated in Figure~\ref{fig:visual}.

\subsection{Other approaches}
\textbf{ProFusion} The main contribution of the Profusion~\citep{profusion} sampling method is a novel technique to ensure the concept's preservation combined with Mixed Sampling. A sampling step in this approach consists of the following stages: (1) we predict $x_t \rightarrow \tilde{x}_{t-1}$ through the usual diffusion backward sampling process with concept (2) after that we make a forward diffusion step $\tilde{x}_{t-1}\rightarrow \tilde{x_t}$ (3) finally, we again make a backward step with the Mixed sampling  $\tilde{x_{t}} \rightarrow x_{t-1}$. The first two steps define Fusion Step and have a hyperparameter $r$ that controls its intensity (e.g. the influence on the result). In case $r=0$ we get Mixed sampling.

\textbf{Photoswap} In this method, the author proposes to replace self-attention features, cross-attention maps, and self-attention maps in the concept trajectory with maps from the superclass at several initial steps. Thus, the method has three hyperparameters: (1) $t_{SF}$ the number of initial steps during which the self-attention features are replaced, (2) $t_{CM}$ the same parameter for cross-attention maps, and (3) $t_{SM}$ for self-attention maps.

\subsection{Evaluation protocol for sampling techniques}
The study of sampling methods involves several key steps. 


The first step is to select a fundamental fine-tuned model that will be used as a baseline for comparing different sampling techniques. For each model, we propose to construct the full Pareto front of Mixed sampling, which we selected as our baseline because it is the simplest yet efficient method, defined by a single hyperparameter.


It is crucial to select a model whose Pareto frontier is of sufficient length, enabling a clearer distinction between the varying parameters. Additionally, this frontier should lie within the optimal balance between concept fidelity and editability compared to other fine-tuning methods. This ensures that we can study sampling in scenarios where the model performs poorly while also confirming that it does not degrade performance when the model excels.

\begin{figure*}[ht!]
\centering
\begin{minipage}{.477\textwidth}
  \centering
  \includegraphics[trim={3cm 10cm 3cm 10cm},clip,width=\linewidth]{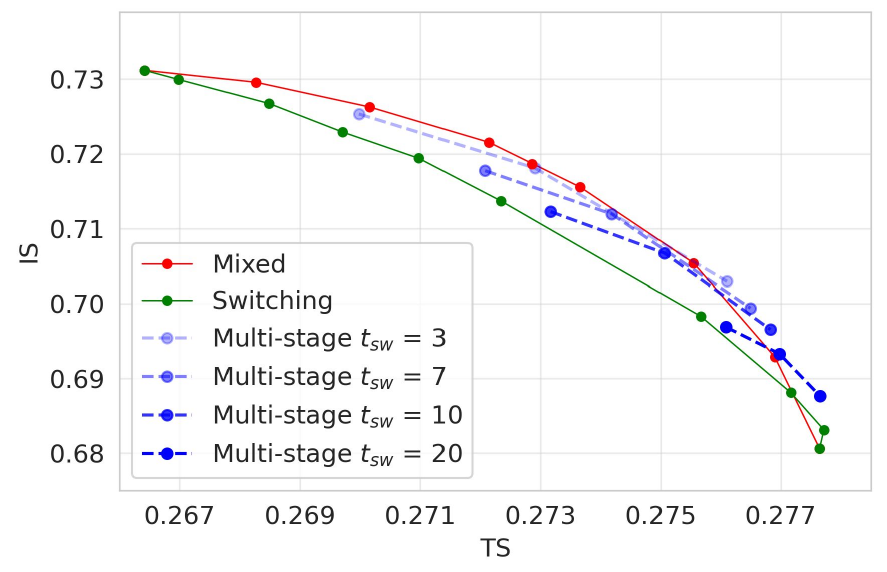}
  \vspace{-0.19in}
  \captionof{figure}{Pareto Frontier curves for Mixed, Switching and Multi-stage Sampling methods. Each Multi-stage sampling curve is generated by fixing the switching step while varying the superclass guidance scale $\omega_s = [1.0, 3.0, 5.0]$.}
  \label{fig:multi-stage}
\end{minipage}%
\hfill
\begin{minipage}{.477\textwidth}
  \centering
  \includegraphics[trim={3cm 10cm 3cm 10cm},clip,width=\linewidth]{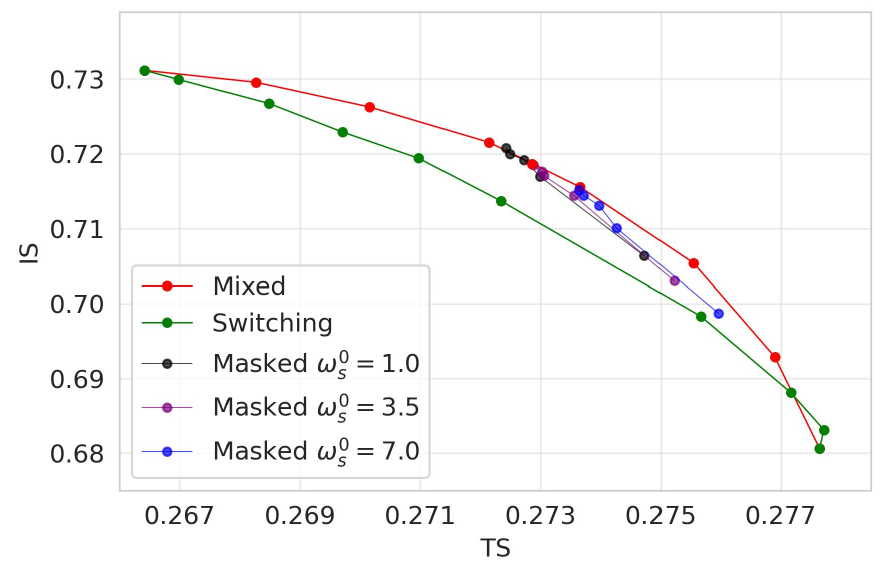}
  \vspace{-0.19in}
  \captionof{figure}{Pareto frontiers curves for  Masked sampling. Each Masked sampling curve is derived by varying the quantile \( q = [ 0.3, 0.5, 0.7, 0.9 ] \), which controls the mask binarization threshold; \( t_{sw} = 3, \omega_s = 3.5\) are fixed.}
  \label{fig:masked}
\end{minipage}
\vspace{-0.19in}
\end{figure*}

\begin{figure}[h!]
    \includegraphics[trim={3cm 10cm 3cm 10cm},clip,width=\linewidth]{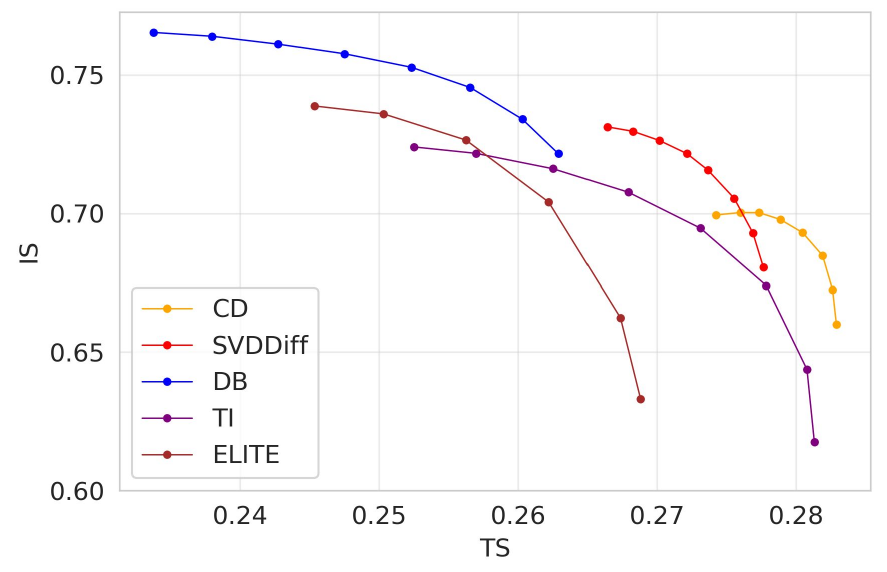}
  \vspace{-0.23in}
    \caption{Mixed sampling Pareto frontiers for different fine-tuning methods.}
    \label{fig:all_mixed}
    \vspace{-0.24in}
\end{figure}
Once the base model is chosen, we fix it and proceed to compare different sampling techniques. For each method, we demonstrate its behaviour at different hyperparameter values. We illustrate the optimal points with generation examples and prove our findings with a user study.

It is important to note that the choice of sampling that maximizes editability can be approached in different ways. For example, one option is to use the model weights before fine-tuning, $\theta^{\text{orig}}$, in (\ref{eq:superclass_sampling}) instead of the fine-tuned weights, $\theta$. Additionally, we can vary the superclass prompts. One extreme option is to remove the superclass token entirely, allowing the model to focus solely on the scene's context (e.g., $p^{\hat{S}} = \textit{"with a city in the background"}$). These hyperparameters affect all sampling methods simultaneously. We analyze this dependency in Appendix~\ref{app:hyper_theta}.
\section{Experiments} \label{sec:experiments}


\textbf{Dataset} We use the Dreambooth~\citep{DB} dataset for evaluation, which contains $30$ concepts from various categories such as pets, interior decoration, toys, and backpacks. To ensure consistency across models, we standardized the data preprocessing procedure (see Appendix~\ref{app:data}). For each concept, we used $25$ contextual text prompts, which include accessorization, appearance, and background modification. For each concept, we generated $10$ images per prompt. In total, there are $750$ unique concept-prompt pairs and a total of $7500$ images for robust evaluation. Additionally, in Appendix~\ref{app:long_prompts}, we compared methods in challenging contexts using 10 long and complicated prompts.

\textbf{Evaluation Metrics}
To estimate the concept's identity preservation, we use CLIP Image Similarity (IS) between real and generated images as in ~\citep{DB}. Higher values for this metric typically indicate better subject fidelity. We also provide DINO for concept fidelity in Appendix~\ref{app:add_dino}. However, it is important to note that as the generated images align more with the contextual prompt, they tend to resemble the original images less. So, even if the identity of the concept is preserved perfectly, the metric will be lower. To evaluate the alignment between generated images and contextual prompts we calculate the CLIP Text Similarity (TS) of the prompt and generated images~\citep{TI}.

\textbf{Selecting base fine-tuning model} At the initial stage, it is essential to choose a foundational fine-tuning model to facilitate the comparison of different sampling methods. To this end, we train five distinct models for each concept, implementing diverse fine-tuning parameterizations~\citep{DB, svdiff}, optimizing text embeddings~\citep{TI, CD}, and leveraging a pre-trained hypernetwork~\citep{elite}. A comprehensive description of the model training and inference procedures can be found in Appendix~\ref{sec:training-details}. We primarily focus on the Stable Diffusion-2-base model in our experiments, but we also include results from other backbones (SD-XL, PixArt-alpha) to further validate our findings in Appendix~\ref{app:add_backbones}.

\begin{figure*}[ht!]
  \centering
  \includegraphics[trim={0 4cm 0 4cm},clip,width=\linewidth]{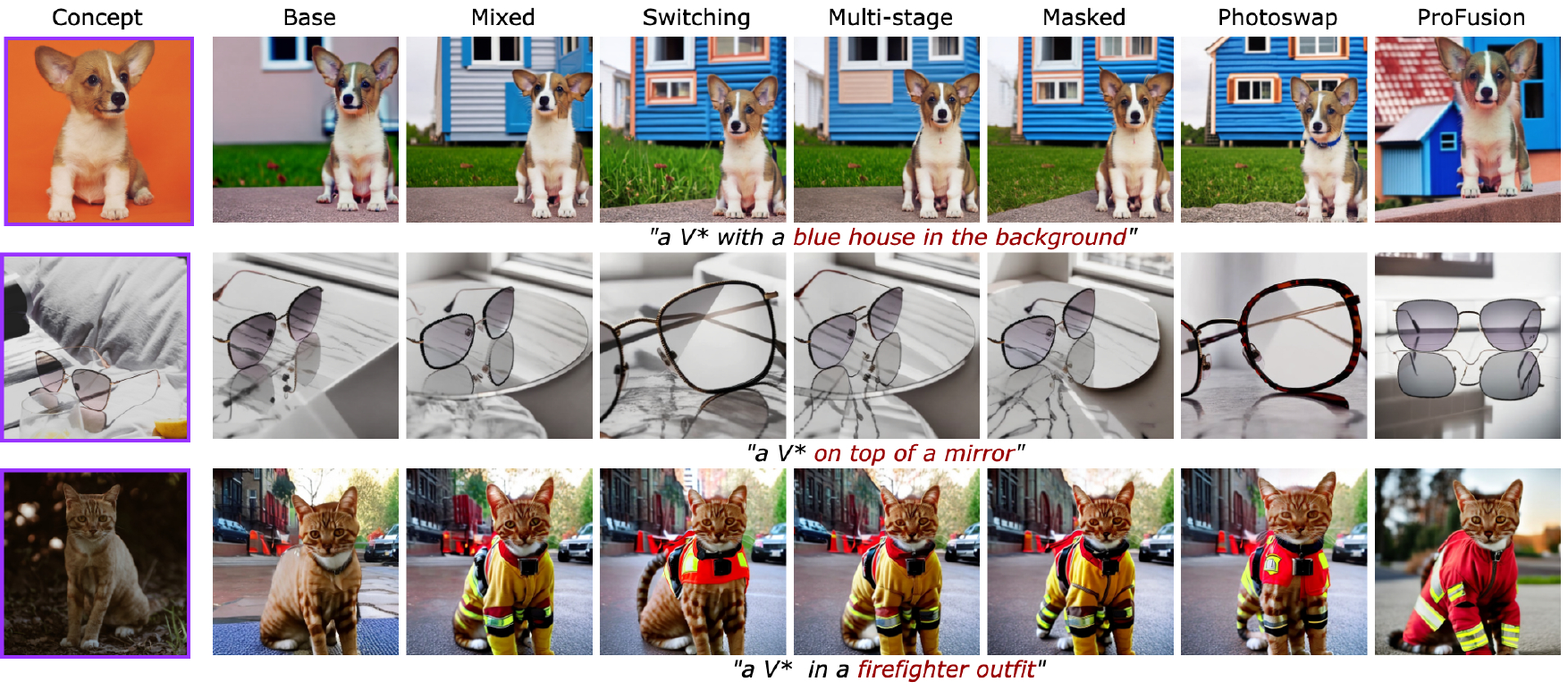}
  \vspace{-0.20in}
  \caption{Examples of the generation outputs for different sampling methods.}
  \label{fig:examples}
  \vspace{-0.21in}
\end{figure*}
For each model, we conduct a complete evaluation of the Mixed sampling by varying the parameter \(\omega_s\) within the range of 0 to 7.0, while deriving \(\omega_c\) as \(7.0 - \omega_s\). Figure~\ref{fig:all_mixed} illustrates the Mixed sampling Pareto frontiers for all aforementioned methods. The method shows the expected behaviour as the superclass guidance scale increases the text similarity improves as well, but the more diverse generation we get, the more we lose on the image similarity. Notably, the results indicate that the Mixed sampling method significantly enhances text similarity across all models. Furthermore, for each model, it is feasible to select a value for \(\omega_s\) such that image similarity remains relatively unchanged, while text similarity is markedly improved.

The Pareto frontier obtained from the SVDiff model achieves a favorable balance between text and image similarity; therefore, this model was chosen for subsequent evaluations of various sampling methods. We complement our analysis with results for different samplings on top of Dreambooth in Appendix~\ref{app:dreambooth}.

\textbf{Computational efficiency of sampling methods}
Switching sampling requires the same number of U-Net calls as Base sampling. Mixed, Multi-stage, Masked, and Photoswap methods double the computational cost by performing two U-Net inferences per step to integrate concept and superclass trajectories. ProFusion further quadruples the number of U-Net inferences per step compared to Base sampling.

\textbf{Proposed sampling techniques analysis}
In Figure~\ref{fig:multi-stage}, the Pareto frontiers for Mixed and Switching sampling are shown. For the Switching sampling, the curve is generated by varying the switching step $t_{sw} = [1, 3, 5, 7, 10, 20, 30, 40]$. We observe that Switching sampling curves lie below the Mixed sampling curve and have lower values of image similarity. This indicates that Switching sampling more negatively affects concept identity.

In addition, we evaluated Multi-stage sampling with various hyperparameters. In Figure~\ref{fig:multi-stage}, each Multi-stage sampling curve is generated by fixing the switching step while varying the superclass guidance scale $\omega_s = [1.0, 3.0, 5.0]$. The graphs show that the curves for Multi-stage sampling fall between the Mixed and Switching Pareto Frontiers. Only curves with high values of the switching step cross the Pareto frontier of Mixed sampling; however, these points correspond to very low image similarity values, thereby compromising concept identity.

Figure~\ref{fig:masked} presents the Pareto frontier for different Masked sampling hyperparameters. For all curves, we fixed the following hyperparameters: \( t_{sw} = 3, \omega_s = 3.5, \omega_c = 3.5 \), as these parameters correspond to the optimal point for Multi-stage and Mixed samplings. Each curve for Masked sampling is derived by varying the quantile \( q = [ 0.3, 0.5, 0.7, 0.9 ] \), which controls the mask binarization threshold. Mask sampling curve falls between the Mixed and Switching Pareto Frontiers and does not facilitate an increase in concept fidelity while maintaining a high quality of alignment with the text. This limitation can be explained by the noisiness of the cross-attention masks, especially during the early stages of generation (see Appendix~\ref{app:cross_attn}), which hampers the precise distinction of the concept. Furthermore, this method is overparameterized, making its practical application challenging.

\begin{figure*}[!ht]
\centering
\begin{minipage}{.477\textwidth}
  \centering
  \includegraphics[trim={3cm 10cm 3cm 10cm},clip,width=\linewidth]{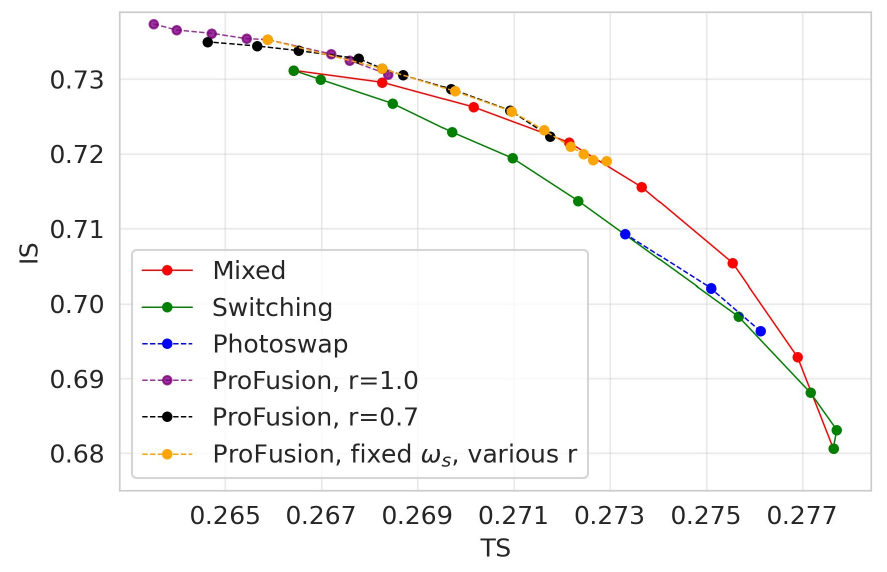}
  \vspace{-0.20in}
  \captionof{figure}{Pareto frontiers curves for Photoswap~\citep{photoswap} and ProFusion~\citep{profusion}.}
  \label{fig:profusion-photoswap}
\end{minipage}
\hfill
\begin{minipage}{.477\textwidth}
  \centering
  \includegraphics[trim={3cm 10cm 3cm 10cm},clip,width=\linewidth]{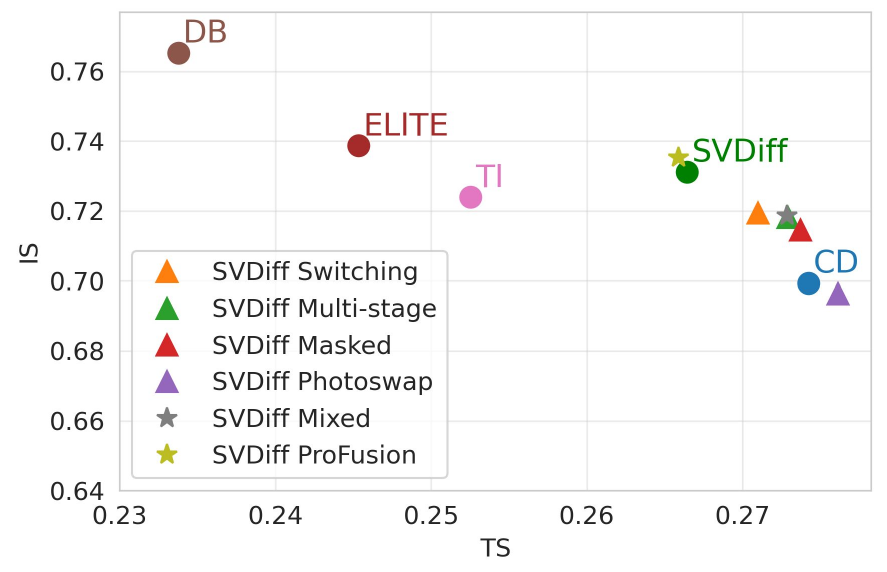}
  \vspace{-0.20in}
  \captionof{figure}{The overall results of different sampling methods against main personalized generation baselines.}
  \label{fig:all-methods}
\end{minipage}
  \vspace{-0.16in}
\end{figure*} 

\textbf{Comparison with existing sampling methods}
To fairly compare our results with Photoswap~\citep{photoswap} and ProFusion~\citep{profusion}, which were initially proposed alongside fixed fine-tuning methods, we reimplemented both approaches using the same fixed SVDiff models to eliminate any influence from differing training methods.

We will first discuss the Photoswap method. In Figure~\ref{fig:profusion-photoswap}, the Pareto front for this method is illustrated. This curve was obtained by varying three hyperparameters: ($t_{SF}, t_{CM}, t_{SM}$) = [(1, 10, 15), (5, 15, 20), (10, 20, 25)], with the last combination representing the optimal values proposed in the original work~\citep{photoswap}. 

As shown in Figure~\ref{fig:profusion-photoswap}, this method's curve is nearly identical to that of Switching sampling. This leads us to conclude that altering the self and cross-attention maps across all layers of the U-Net affects generation almost equally as using the entire noise prediction from the superclass trajectory.

Additionally, the ProFusion Pareto frontiers are shown in Figure~\ref{fig:profusion-photoswap}. Since Mixed sampling is part of the ProFusion, we evaluated it in the same way by fixing all parameters and varying $\omega_s$. We assessed this method using two levels of fusion step intensity $r$ and constructed a distinct curve with fixed $\omega_s=3.5$ and various $r = [0.05, 0.1, 0.15, 0.2, 0.3, 0.4, 0.5, 0.7, 1.0]$. As observed, with decreasing fusion step intensity $r$, the curve converges more closely to the Mixed sampling curve. However, when the fusion step intensity is high, this method significantly improves concept preservation and results in image similarity even higher than Base sampling.

\begin{figure*}[t!]
  \centering
  \includegraphics[trim={0 4cm 0 4cm},clip,width=\linewidth]{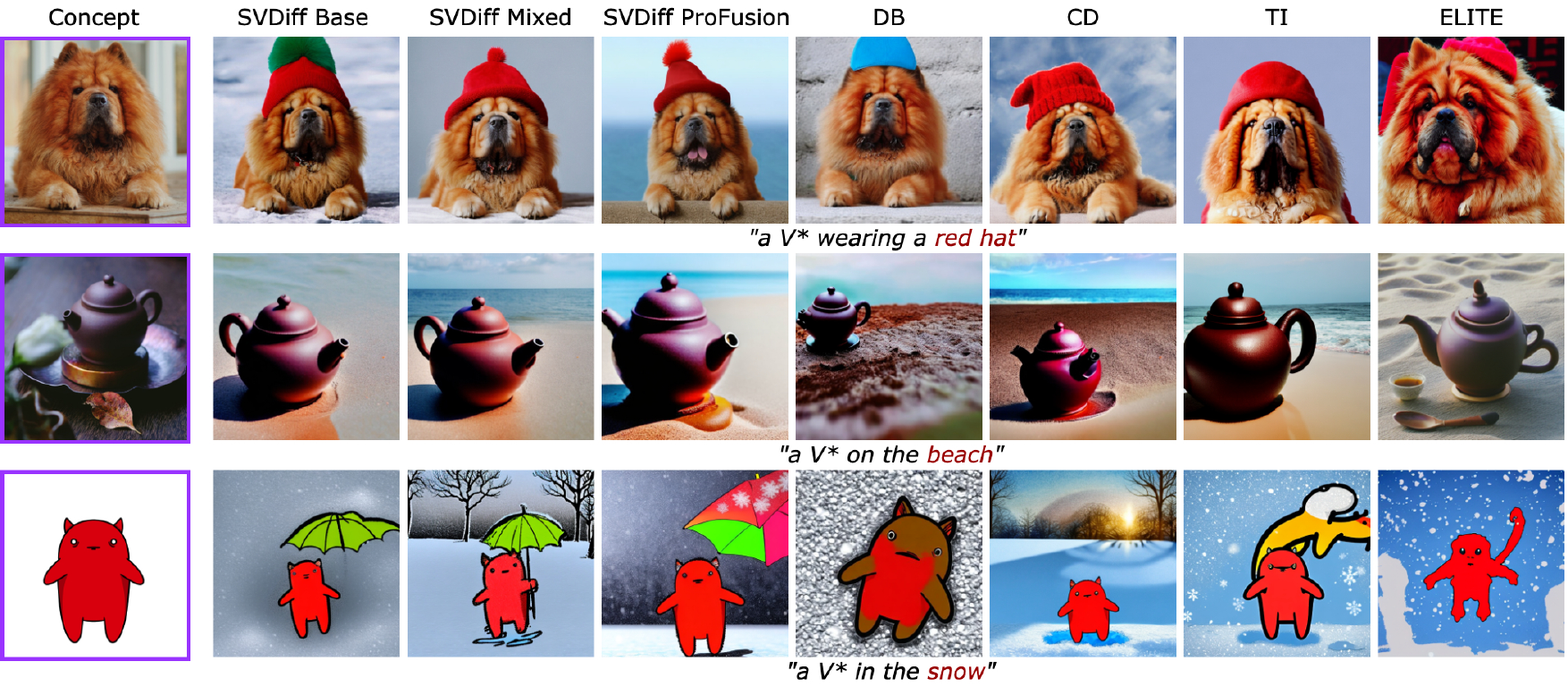}
  \caption{Examples of generation results for Mixed and ProFusion sampling methods compared to the main personalized generation baselines.}
  \label{fig:main_examples}
\end{figure*}

\textbf{User study}
In addition to CLIP metrics, we also conducted a human evaluation. For each sampling method, we took the optimal point in terms of CLIP metric and visual generation assessment and generated \num{16000} pairs comparing different sampling techniques and base personalization methods (Dreambooth (DB), Custom Diffusion (CD), Textual Inversion (TI) and ELITE) with Mixed Sampling as a strong and effective baseline. See Appendix~\ref{app:us} for more details.

Given an original image of the concept, a text prompt, and $2$ generated images (Mixed versus the competitor's), we asked users to answer the following questions: 1) "Which image is more consistent with the text prompt?" to evaluate text similarity 2) "Which image better represents the original image?" for image similarity 3) "Which image is generally better in terms of alignment with the prompt and concept identity preservation?" to evaluate the general impression. We provide an example of a comparison in Appendix~\ref{app:us}.

\begin{table*}[ht!]
\centering
\caption{User study results of the pairwise comparison of SVDDiff with Mixed sampling method versus other baselines.
The values in the table show the win rate. "TS" stands for text similarity, "IS" stands for image similarity, and "All" stands for overall impression. *Sampling method is applied on top of the SVDDiff fine-tuning method.}
	\label{table:user_study}
\resizebox{\textwidth}{!}{

\begin{tabular}{lccccccccccccccc}
\toprule
 & Base* & Switching* & Multi-stage* & Masked* & Photoswap* & ProFusion*  & DB & TI & ELITE & CD \\
\midrule
TS & $0.52$ & $0.51$ & $0.51$ & $0.51$ & $0.53$ & $0.49$ & $0.74$ & $0.67$ & $0.64$ & $0.51$ \\
IS & $0.37$ & $0.47$ & $0.50$ & $0.59$ & $0.70$ & $0.35$ & $0.40$ & $0.74$ & $0.73$ & $0.53$ \\
All & $0.41$ & $0.48$ & $0.50$ & $0.59$ & $0.69$ & $0.37$ & $0.59$ & $0.77$ & $0.75$ & $0.53$ \\
\bottomrule
\end{tabular}
}
\end{table*}

Combining the user study results (Table~\ref{table:user_study}) and the insights from Figure~\ref{fig:all-methods}, which illustrates the improvements of the examined techniques against the main personalized generation baselines, we find that all sampling methods improve the performance of the fine-tuned model in either concept or context preservation.






\textbf{A framework for selecting sampling method}
In this section, we analyze the performance of different sampling methods in terms of concept fidelity, alignment with a text prompt, and computational efficiency. Our conclusions are based primarily on the user study results, as research suggests that CLIP metrics do not always align with human perception. When the user study does not reveal significant differences between methods, we rely on quantitative metrics and visual examples to guide our recommendations.

\textbf{Step 1: Assess Text Alignment Needs}
If Base sampling fails to align adequately with the text prompt — as illustrated in Figure~\ref{fig:examples} — consider alternative methods that improve text similarity. Mixed, Switching, Multi-stage, and Masked sampling all achieve comparable text similarity, as demonstrated by both user study results and CLIP metrics.

\textbf{Step 2: Evaluate Computational Efficiency} Switching sampling is the simplest and most cost-effective option, improving TS without increasing computational load, though it may compromise concept preservation. Mixed sampling, requiring twice as many U-Net inferences of Base or Switching, offers more stable results with better concept and context preservation (see Figures~\ref{fig:examples},~\ref{fig:add_ex}).

\textbf{Step 3: Prioritize Concept Fidelity} Switching, Mixed, and Masked samplings may struggle to achieve better concept fidelity. Meanwhile, ProFusion excels in both text similarity and concept preservation, as supported by user feedback (Figures~\ref{fig:main_examples} and~\ref{fig:add_ex_all}). The trade-off is computational cost: ProFusion requires four times more U-Net inference than Base sampling and requires careful hyperparameter selection.

\textbf{Step 4: Make an Informed Decision}

\begin{itemize}
    \item \textbf{Switching sampling} matches the computational load of Base sampling, improving text similarity at the expense of concept fidelity.
    \item \textbf{Mixed sampling} requires twice as many U-Net inferences of Base sampling and outperforms Switching in balancing text similarity and concept preservation.
    \item \textbf{ProFusion} requires four times the U-Net inferences of Base sampling, achieving the highest concept fidelity but requiring extensive hyperparameter tuning and offering less text similarity improvement than Mixed.
\end{itemize}

\section{Conclusion}
\label{sec:conclusion}

In this paper, we investigate the role of sampling methods in improving personalized text-to-image generation, focusing on their interaction with fine-tuning strategies and their impact on concept fidelity and adaptability. Using systematic evaluations, we demonstrate that integrating superclass trajectories into the sampling process leads to significant improvements, offering a flexible approach to balancing concept preservation and prompt adherence across diverse architectures. Our analysis provides a comprehensive framework for understanding the trade-offs between different sampling techniques and their application in a variety of generative scenarios. We hope that this study will inspire further research on decoupling fine-tuning from sampling for improvement and exploring the potential of these methods independently.

Regarding the limitations of sampling techniques, we highlight two main issues. First, the sampling methods require careful hyperparameter tuning, and finding the optimal configuration for each technique can be challenging. Second, some of the more advanced sampling techniques, such as ProFusion, require higher computational costs, making them less practical for real-time or large-scale applications compared to simpler alternatives.

\clearpage

\section*{Impact Statement}
Our work advances personalized text-to-image generation by systematically optimizing sampling strategies to balance concept fidelity and adaptability, enabling the efficient creation of tailored visual content for applications in creative design, education, and accessibility tools. 

Potential harms include misuse for generating deceptive or harmful content, such as misinformation, non-consensual deepfakes, or targeted impersonations, raising ethical concerns around privacy, consent, and trust in digital media. We emphasize the need for safeguards, transparency, and responsible deployment to mitigate these risks while fostering beneficial innovation.

\bibliography{beyond_ft_paper}

\begin{thebibliography}{23}
\providecommand{\natexlab}[1]{#1}
\providecommand{\url}[1]{\texttt{#1}}
\expandafter\ifx\csname urlstyle\endcsname\relax
  \providecommand{\doi}[1]{doi: #1}\else
  \providecommand{\doi}{doi: \begingroup \urlstyle{rm}\Url}\fi

\bibitem[Chen et~al.(2023{\natexlab{a}})Chen, Zhang, Wu, Wang, Duan, Zhou, and
  Zhu]{disenbooth}
Chen, H., Zhang, Y., Wu, S., Wang, X., Duan, X., Zhou, Y., and Zhu, W.
\newblock Disenbooth: Identity-preserving disentangled tuning for
  subject-driven text-to-image generation.
\newblock In \emph{The Twelfth International Conference on Learning
  Representations}, 2023{\natexlab{a}}.

\bibitem[Chen et~al.(2023{\natexlab{b}})Chen, Yu, Ge, Yao, Xie, Wu, Wang, Kwok,
  Luo, Lu, and Li]{chen2023pixartalphafasttrainingdiffusion}
Chen, J., Yu, J., Ge, C., Yao, L., Xie, E., Wu, Y., Wang, Z., Kwok, J., Luo,
  P., Lu, H., and Li, Z.
\newblock Pixart-$\alpha$: Fast training of diffusion transformer for
  photorealistic text-to-image synthesis, 2023{\natexlab{b}}.
\newblock URL \url{https://arxiv.org/abs/2310.00426}.

\bibitem[Gal et~al.(2022)Gal, Alaluf, Atzmon, Patashnik, Bermano, Chechik, and
  Cohen-Or]{TI}
Gal, R., Alaluf, Y., Atzmon, Y., Patashnik, O., Bermano, A.~H., Chechik, G.,
  and Cohen-Or, D.
\newblock An image is worth one word: Personalizing text-to-image generation
  using textual inversion.
\newblock \emph{arXiv preprint arXiv:2208.01618}, 2022.

\bibitem[Gu et~al.(2024)Gu, Wang, Zhao, Fu, Xiong, Liu, Zhang, Zhang, Zhang,
  Jung, et~al.]{photoswap}
Gu, J., Wang, Y., Zhao, N., Fu, T.-J., Xiong, W., Liu, Q., Zhang, Z., Zhang,
  H., Zhang, J., Jung, H., et~al.
\newblock Photoswap: Personalized subject swapping in images.
\newblock \emph{Advances in Neural Information Processing Systems}, 36, 2024.

\bibitem[Han et~al.(2023)Han, Li, Zhang, Milanfar, Metaxas, and Yang]{svdiff}
Han, L., Li, Y., Zhang, H., Milanfar, P., Metaxas, D., and Yang, F.
\newblock Svdiff: Compact parameter space for diffusion fine-tuning.
\newblock In \emph{Proceedings of the IEEE/CVF International Conference on
  Computer Vision}, pp.\  7323--7334, 2023.

\bibitem[Ho \& Salimans(2022)Ho and Salimans]{classfree}
Ho, J. and Salimans, T.
\newblock Classifier-free diffusion guidance.
\newblock \emph{arXiv preprint arXiv:2207.12598}, 2022.

\bibitem[Hu et~al.(2021)Hu, Shen, Wallis, Allen-Zhu, Li, Wang, Wang, and
  Chen]{lora}
Hu, E.~J., Shen, Y., Wallis, P., Allen-Zhu, Z., Li, Y., Wang, S., Wang, L., and
  Chen, W.
\newblock Lora: Low-rank adaptation of large language models.
\newblock \emph{arXiv preprint arXiv:2106.09685}, 2021.

\bibitem[Kumari et~al.(2023)Kumari, Zhang, Zhang, Shechtman, and Zhu]{CD}
Kumari, N., Zhang, B., Zhang, R., Shechtman, E., and Zhu, J.-Y.
\newblock Multi-concept customization of text-to-image diffusion.
\newblock In \emph{Proceedings of the IEEE/CVF Conference on Computer Vision
  and Pattern Recognition}, pp.\  1931--1941, 2023.

\bibitem[Oquab et~al.(2024)Oquab, Darcet, Moutakanni, Vo, Szafraniec, Khalidov,
  Fernandez, Haziza, Massa, El-Nouby, Assran, Ballas, Galuba, Howes, Huang, Li,
  Misra, Rabbat, Sharma, Synnaeve, Xu, Jegou, Mairal, Labatut, Joulin, and
  Bojanowski]{oquab2024dinov2learningrobustvisual}
Oquab, M., Darcet, T., Moutakanni, T., Vo, H., Szafraniec, M., Khalidov, V.,
  Fernandez, P., Haziza, D., Massa, F., El-Nouby, A., Assran, M., Ballas, N.,
  Galuba, W., Howes, R., Huang, P.-Y., Li, S.-W., Misra, I., Rabbat, M.,
  Sharma, V., Synnaeve, G., Xu, H., Jegou, H., Mairal, J., Labatut, P., Joulin,
  A., and Bojanowski, P.
\newblock Dinov2: Learning robust visual features without supervision, 2024.
\newblock URL \url{https://arxiv.org/abs/2304.07193}.

\bibitem[Podell et~al.(2023)Podell, English, Lacey, Blattmann, Dockhorn,
  Müller, Penna, and Rombach]{podell2023sdxlimprovinglatentdiffusion}
Podell, D., English, Z., Lacey, K., Blattmann, A., Dockhorn, T., Müller, J.,
  Penna, J., and Rombach, R.
\newblock Sdxl: Improving latent diffusion models for high-resolution image
  synthesis, 2023.
\newblock URL \url{https://arxiv.org/abs/2307.01952}.

\bibitem[Qiu et~al.(2024)Qiu, Liu, Feng, Xue, Feng, Liu, Zhang, Weller, and
  Sch{\"o}lkopf]{ortogonal}
Qiu, Z., Liu, W., Feng, H., Xue, Y., Feng, Y., Liu, Z., Zhang, D., Weller, A.,
  and Sch{\"o}lkopf, B.
\newblock Controlling text-to-image diffusion by orthogonal finetuning.
\newblock \emph{Advances in Neural Information Processing Systems}, 36, 2024.

\bibitem[Radford et~al.(2021)Radford, Kim, Hallacy, Ramesh, Goh, Agarwal,
  Sastry, Askell, Mishkin, Clark, et~al.]{clip}
Radford, A., Kim, J.~W., Hallacy, C., Ramesh, A., Goh, G., Agarwal, S., Sastry,
  G., Askell, A., Mishkin, P., Clark, J., et~al.
\newblock Learning transferable visual models from natural language
  supervision.
\newblock In \emph{International conference on machine learning}, pp.\
  8748--8763. PMLR, 2021.

\bibitem[Ramesh et~al.(2021)Ramesh, Pavlov, Goh, Gray, Voss, Radford, Chen, and
  Sutskever]{ramesh2021zero}
Ramesh, A., Pavlov, M., Goh, G., Gray, S., Voss, C., Radford, A., Chen, M., and
  Sutskever, I.
\newblock Zero-shot text-to-image generation.
\newblock In \emph{International conference on machine learning}, pp.\
  8821--8831. Pmlr, 2021.

\bibitem[Ramesh et~al.(2022)Ramesh, Dhariwal, Nichol, Chu, and
  Chen]{ramesh2022hierarchical}
Ramesh, A., Dhariwal, P., Nichol, A., Chu, C., and Chen, M.
\newblock Hierarchical text-conditional image generation with clip latents.
\newblock \emph{arXiv preprint arXiv:2204.06125}, 1\penalty0 (2):\penalty0 3,
  2022.

\bibitem[Rombach et~al.(2022{\natexlab{a}})Rombach, Blattmann, Lorenz, Esser,
  and Ommer]{rombach2022high}
Rombach, R., Blattmann, A., Lorenz, D., Esser, P., and Ommer, B.
\newblock High-resolution image synthesis with latent diffusion models.
\newblock In \emph{Proceedings of the IEEE/CVF conference on computer vision
  and pattern recognition}, pp.\  10684--10695, 2022{\natexlab{a}}.

\bibitem[Rombach et~al.(2022{\natexlab{b}})Rombach, Blattmann, Lorenz, Esser,
  and Ommer]{stablediffusion}
Rombach, R., Blattmann, A., Lorenz, D., Esser, P., and Ommer, B.
\newblock High-resolution image synthesis with latent diffusion models.
\newblock In \emph{Proceedings of the IEEE/CVF conference on computer vision
  and pattern recognition}, pp.\  10684--10695, 2022{\natexlab{b}}.

\bibitem[Ronneberger et~al.(2015)Ronneberger, Fischer, and Brox]{unet}
Ronneberger, O., Fischer, P., and Brox, T.
\newblock U-net: Convolutional networks for biomedical image segmentation.
\newblock In \emph{Medical image computing and computer-assisted
  intervention--MICCAI 2015: 18th international conference, Munich, Germany,
  October 5-9, 2015, proceedings, part III 18}, pp.\  234--241. Springer, 2015.

\bibitem[Ruiz et~al.(2023)Ruiz, Li, Jampani, Pritch, Rubinstein, and
  Aberman]{DB}
Ruiz, N., Li, Y., Jampani, V., Pritch, Y., Rubinstein, M., and Aberman, K.
\newblock Dreambooth: Fine tuning text-to-image diffusion models for
  subject-driven generation.
\newblock In \emph{Proceedings of the IEEE/CVF Conference on Computer Vision
  and Pattern Recognition}, pp.\  22500--22510, 2023.

\bibitem[Saharia et~al.(2022)Saharia, Chan, Saxena, Li, Whang, Denton,
  Ghasemipour, Gontijo~Lopes, Karagol~Ayan, Salimans,
  et~al.]{saharia2022photorealistic}
Saharia, C., Chan, W., Saxena, S., Li, L., Whang, J., Denton, E.~L.,
  Ghasemipour, K., Gontijo~Lopes, R., Karagol~Ayan, B., Salimans, T., et~al.
\newblock Photorealistic text-to-image diffusion models with deep language
  understanding.
\newblock \emph{Advances in neural information processing systems},
  35:\penalty0 36479--36494, 2022.

\bibitem[Song et~al.(2020)Song, Meng, and Ermon]{ddim}
Song, J., Meng, C., and Ermon, S.
\newblock Denoising diffusion implicit models.
\newblock \emph{arXiv preprint arXiv:2010.02502}, 2020.

\bibitem[Tewel et~al.(2023)Tewel, Gal, Chechik, and Atzmon]{r1e}
Tewel, Y., Gal, R., Chechik, G., and Atzmon, Y.
\newblock Key-locked rank one editing for text-to-image personalization.
\newblock In \emph{ACM SIGGRAPH 2023 Conference Proceedings}, pp.\  1--11,
  2023.

\bibitem[Wei et~al.(2023)Wei, Zhang, Ji, Bai, Zhang, and Zuo]{elite}
Wei, Y., Zhang, Y., Ji, Z., Bai, J., Zhang, L., and Zuo, W.
\newblock Elite: Encoding visual concepts into textual embeddings for
  customized text-to-image generation.
\newblock In \emph{Proceedings of the IEEE/CVF International Conference on
  Computer Vision}, pp.\  15943--15953, 2023.

\bibitem[Zhou et~al.(2023)Zhou, Zhang, Sun, and Xu]{profusion}
Zhou, Y., Zhang, R., Sun, T., and Xu, J.
\newblock Enhancing detail preservation for customized text-to-image
  generation: A regularization-free approach.
\newblock \emph{arXiv preprint arXiv:2305.13579}, 2023.

\end{thebibliography}
\bibliographystyle{icml2025}

\newpage
\appendix
\onecolumn

\section{Related Work} \label{app:related_work}
\textbf{Personalized Generation} 
Due to the considerable success of large text-to-image models \cite{ramesh2022hierarchical, ramesh2021zero, saharia2022photorealistic, rombach2022high}, the field of personalized generation has been actively developed. The challenge is to customize a text-to-image model to generate specific concepts that are specified using several input images. Many different approaches \cite{DB, TI, CD, svdiff, ortogonal, profusion, elite, r1e} have been proposed to solve this problem and can be divided into the following groups: pseudo-token optimization \cite{TI, profusion, disenbooth, r1e}, diffusion fune-tuning \cite{DB, CD, profusion}, and encoder-based \cite{elite}. The pseudo-token paradigm adjusts the text encoder to convert the concept token into the proper embedding for the diffusion model. Such embedding can be optimized directly \cite{TI, r1e} or can be generated by other neural networks \cite{disenbooth, profusion}. Such approaches usually require a small number of parameters to optimize but lose the visual features of the target concept. Diffusion fine-tuning-based methods optimize almost all \cite{DB} or parts \cite{CD} of the model to reconstruct the training images of the concept. This allows the model to learn the input concept with high accuracy, but the model due to overfitting may lose the ability to edit it when generated with different text prompts. To reduce overfitting and memory usage, lightweight parameterizations \cite{svdiff, r1e, lora} have been proposed that preserve edibility but at the cost of degrading concept fidelity. Encoder-based methods \cite{elite} allow one forward pass of an encoder that has been trained on a large dataset of many different objects to embed the input concept. This dramatically speeds up the process of learning a new concept and such a model is highly editable, but the quality of recovering concept details may be low. Generally, the main problem with existing personalized generation approaches is that they struggle to simultaneously recover a concept with high quality and generate it in a variety of scenes.

\textbf{Sampling strategies}
Much research has been devoted to sampling techniques for text-to-image diffusion models, focusing not only on personalized generation but also on image editing. In this paper, we address a more specific question: how can the two trajectories -- superclass and concept -- be optimally combined to achieve both high concept fidelity and high editability? The ProFusion paper \cite{profusion} considered one way of combining these trajectories (Mixed sampling), which we analyze in detail in our paper (see Section \ref{sec:mixed_sampling}) and show its properties and problems. In ProFusion, authors additionally proposed a more complex sampling procedure, which we observed to be redundant compared to Mixed sampling, as can be seen in our experiments (see Section \ref{sec:experiments}). In Photoswap \cite{photoswap}, authors consider another way of combining trajectories by superclass and concept, which turns out to be almost identical to the Switching sampling strategy that we discuss in detail in Section \ref{sec:switching_sampling}. We show why this strategy fails to achieve simultaneous improvements in concept reconstruction and editability. In the paper, we propose a more efficient way of combining these two trajectories that achieves an optimal balance between the two key features of personalized generation: concept reconstruction and editability.

\section{Training details} \label{sec:training-details}
The Stable Diffusion-2-base model is used for all experiments. For the Dreambooth, Custom Diffusion, and Textual Inversion methods, we used the implementation from \url{https://github.com/huggingface/diffusers}.

\textbf{SVDiff} We implement the method based on \url{https://github.com/mkshing/svdiff-pytorch}. The parameterization is applied to all Text Encoder and U-Net layers. The models for all concepts were trained for $1600$ using Adam optimizer with $\text{batch size} = 1$, $\text{learning rate} = 0.001$, $\text{learning rate 1d} = 0.000001$, $\text{betas} = (0.9, 0.999)$, $\text{epsilon} = 1e\!-\!8$, and $\text{weight decay} = 0.01$. 

\textbf{Dreambooth} All query, key, and value layers in Text Encoder and U-Net were trained during fine-tuning. The models for all concepts were trained for $400$ steps using Adam optimizer with $\text{batch size} = 1$, $\text{learning rate} = 2e\!-\!5$, $\text{betas} = (0.9, 0.999)$, $\text{epsilon} = 1e\!-\!8$, and $\text{weight decay} = 0.01$. 

\textbf{Custom Diffusion} The models for all concepts were trained for $1600$ steps using Adam optimizer with $\text{batch size} = 1$, $\text{learning rate} = 0.00001$, $\text{betas} = (0.9, 0.999)$, $\text{epsilon} = 1e\!-\!8$, and $\text{weight decay} = 0.01$. 

\textbf{Textual Inversion} The models for all concepts were trained for $10000$ steps using Adam optimizer with $\text{batch size} = 1$, $\text{learning rate} = 0.005$, $\text{betas} = (0.9, 0.999)$, $\text{epsilon} = 1e\!-\!8$, and $\text{weight decay} = 0.01$. 

\textbf{ELITE} We used the pre-trained model from the official repo \url{https://github.com/csyxwei/ELITE} with $\lambda=0.6$ and inference hyperparams from the original paper.

\clearpage
\section{Superclass and concept trajectory choice}\label{app:hyper_theta}

 \begin{wrapfigure}{r}{0.45\textwidth}
    \includegraphics[trim={3cm 10cm 3cm 10cm},clip,width=\linewidth]{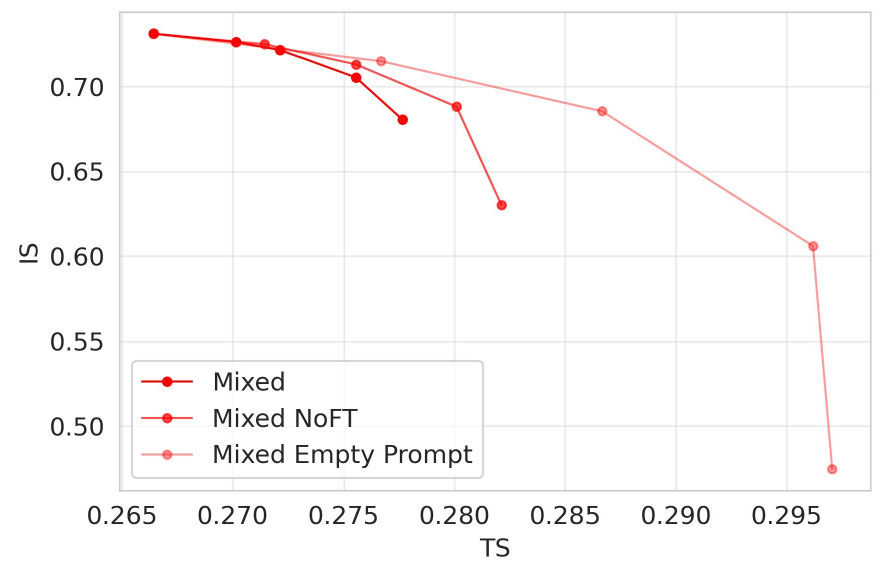}
    \caption{The Pareto frontiers for original Mixed sampling and Mixed sampling in the Superclass, NoFT, and Empty Prompt setups. Mixed NoFT and Mixed Empty Prompt configurations overlap with the Pareto frontier of the original mixed sampling, but primarily in regions associated with low image similarity, which compromises concept fidelity.} \label{fig:mixed_noft_ep}
    \vspace{-0.14in}
\end{wrapfigure}

There are multiple ways to define sampling with maximized textual alignment to the prompt. However, the arbitrary choice can harm the alignment between Base sampling~\ref{eq:concept_sampling} and the selected trajectory. We use the Sampling with superclass (\ref{eq:superclass_sampling}) as it's the default choice in the literature and guarantees the maximized alignment between noise predictions $\tilde{\varepsilon}_{\theta}(p^C)$ and $\tilde{\varepsilon}_{\theta}(p^S)$. 

The several natural ways to adjust Sampling with superclass can be presented by varying $\theta$ and $p^{S}$ in (\ref{eq:superclass_sampling}). We explore two additional options with decreased alignment with (\ref{eq:concept_sampling}): (1) NoFT -- weights of base model $\theta^{\text{orig}}$ instead fine-tuned weights, (2) Empty Prompt -- prompt without any reference to a concept, even to its superclass category, i.e. $p^{\hat{S}} = \textit{"with a city in the background"}$ instead of $p^{S} = \textit{"a backpack with a city in the background"}$.

\begin{figure}[b]
    \centering
    \includegraphics[trim={3cm 10cm 3cm 10cm},clip,width=0.32\linewidth]{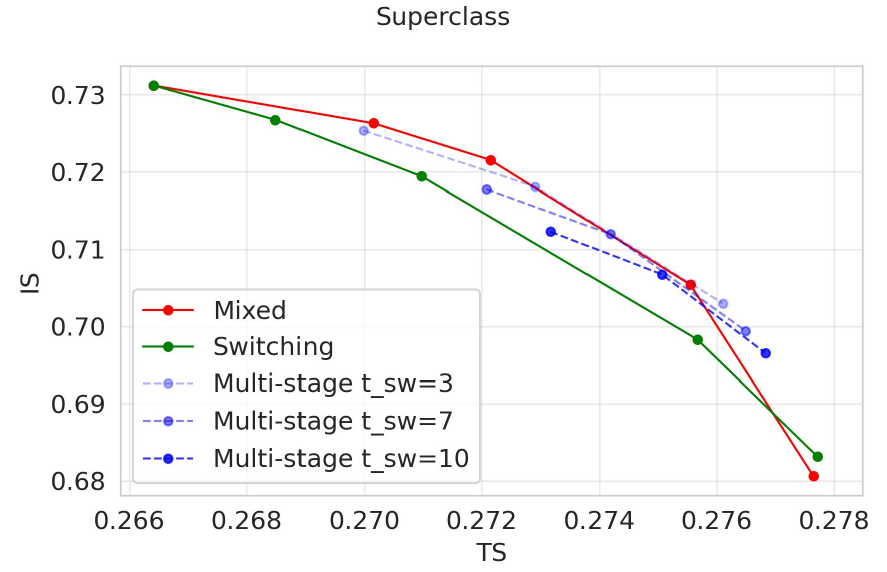}
    \hfill
    \includegraphics[trim={3cm 10cm 3cm 10cm},clip,width=0.32\linewidth]{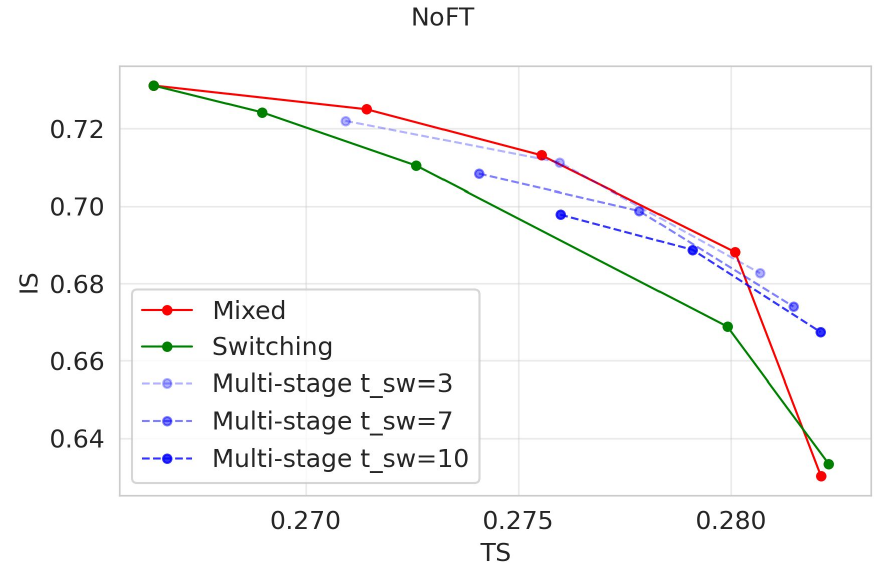}
    \hfill
    \includegraphics[trim={3cm 10cm 3cm 10cm},clip,width=0.32\linewidth]{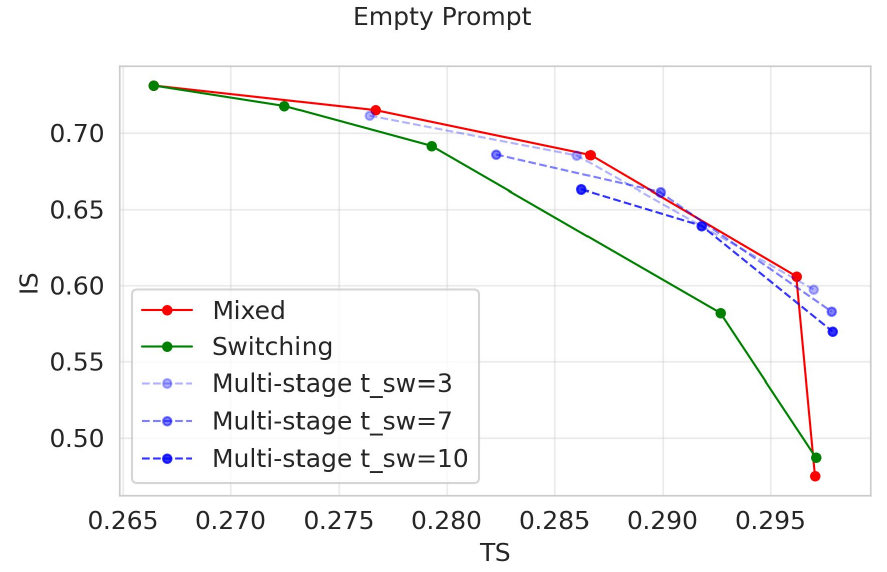}
    \caption{Pareto Frontier Curves for Mixed, Switching, and Multi-Stage Sampling Methods in the Superclass, NoFT and Empty Prompt setups.
The NoFT and Empty Prompt configurations (second and third columns, respectively) preserve the same method ordering as those produced by Superclass sampling (first column).} \label{fig:multi-stage_noft_ep}
\end{figure}

\begin{figure}[t]
    \centering
    \includegraphics[trim={3cm 10cm 3cm 10cm},clip,width=0.32\linewidth]{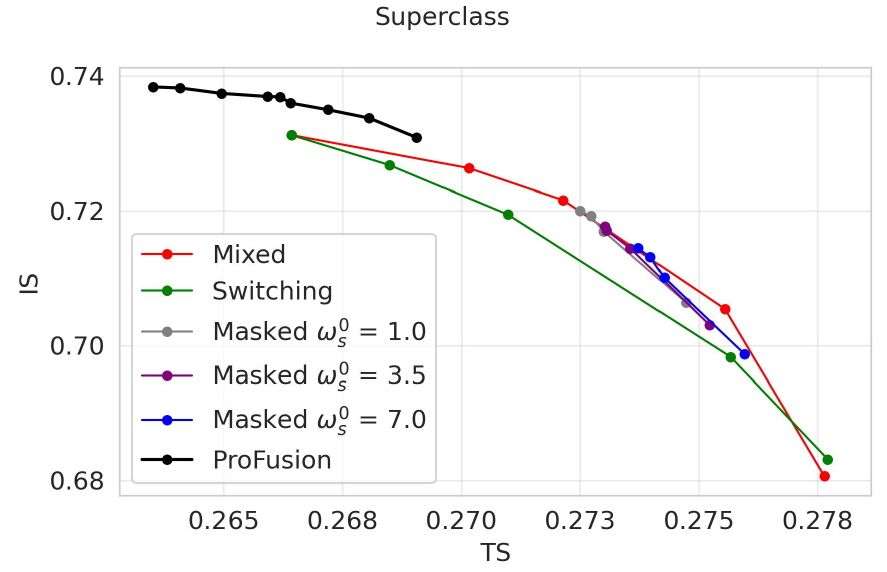}
    \hfill
    \includegraphics[trim={3cm 10cm 3cm 10cm},clip,width=0.32\linewidth]{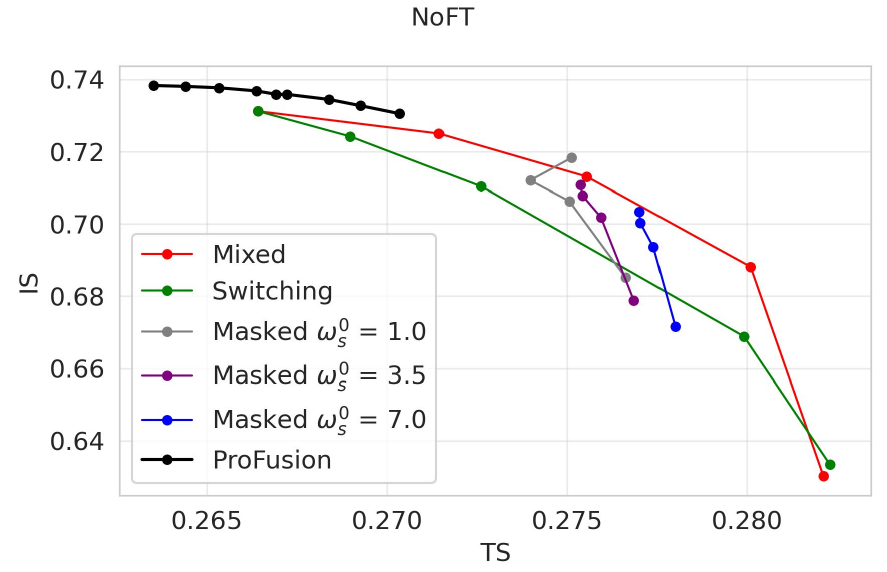}
    \hfill
    \includegraphics[trim={3cm 10cm 3cm 10cm},clip,width=0.32\linewidth]{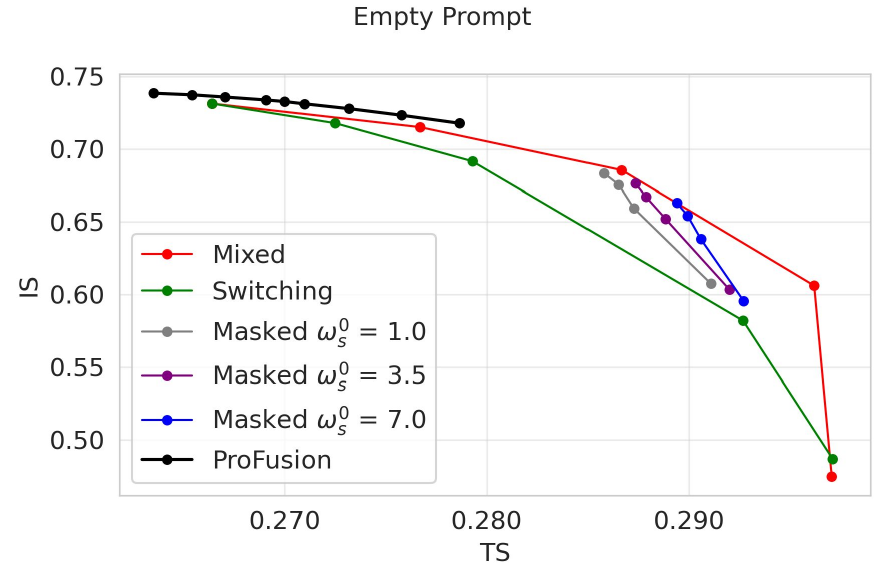}
    \caption{Pareto Frontier Curves for Mixed, Switching, Masked, and ProFusion Sampling Methods in the Superclass, NoFT, and Empty Prompt setups.
The NoFT and Empty Prompt configurations (second and third columns, respectively) preserve the same method ordering as those produced by Superclass sampling (first column).} \label{fig:masked_noft_ep}
\end{figure}

To validate the robustness of our framework for sampling method selection, we employ the original experimental protocol, supplementing the results shown in Figures~\ref{fig:examples} and~\ref{fig:profusion-photoswap}. Our analysis of Figures~\ref{fig:multi-stage_noft_ep} and~\ref{fig:masked_noft_ep} reveals that trajectories generated under the NoFT and Empty Prompt configurations (second and third columns, respectively) maintain identical method ordering to those produced by Superclass sampling ((\ref{eq:superclass_sampling}), first column).

Notably, Figure~\ref{fig:mixed_noft_ep} shows that Empty Prompt configuration demonstrates weaker alignment with Base sampling compared to NoFT, particularly at higher values of the superclass guidance scale $\omega_{s}$. This divergence manifests as reduced concept fidelity for Empty Prompt under large $\omega_{s}$. These findings highlight a practical adjustment: prioritizing smaller $\omega_{s}$ values in Empty Prompt setup preserves concept fidelity without altering the framework’s core selection logic. 

\begin{figure}[b]
    \centering
    \includegraphics[width=\linewidth]{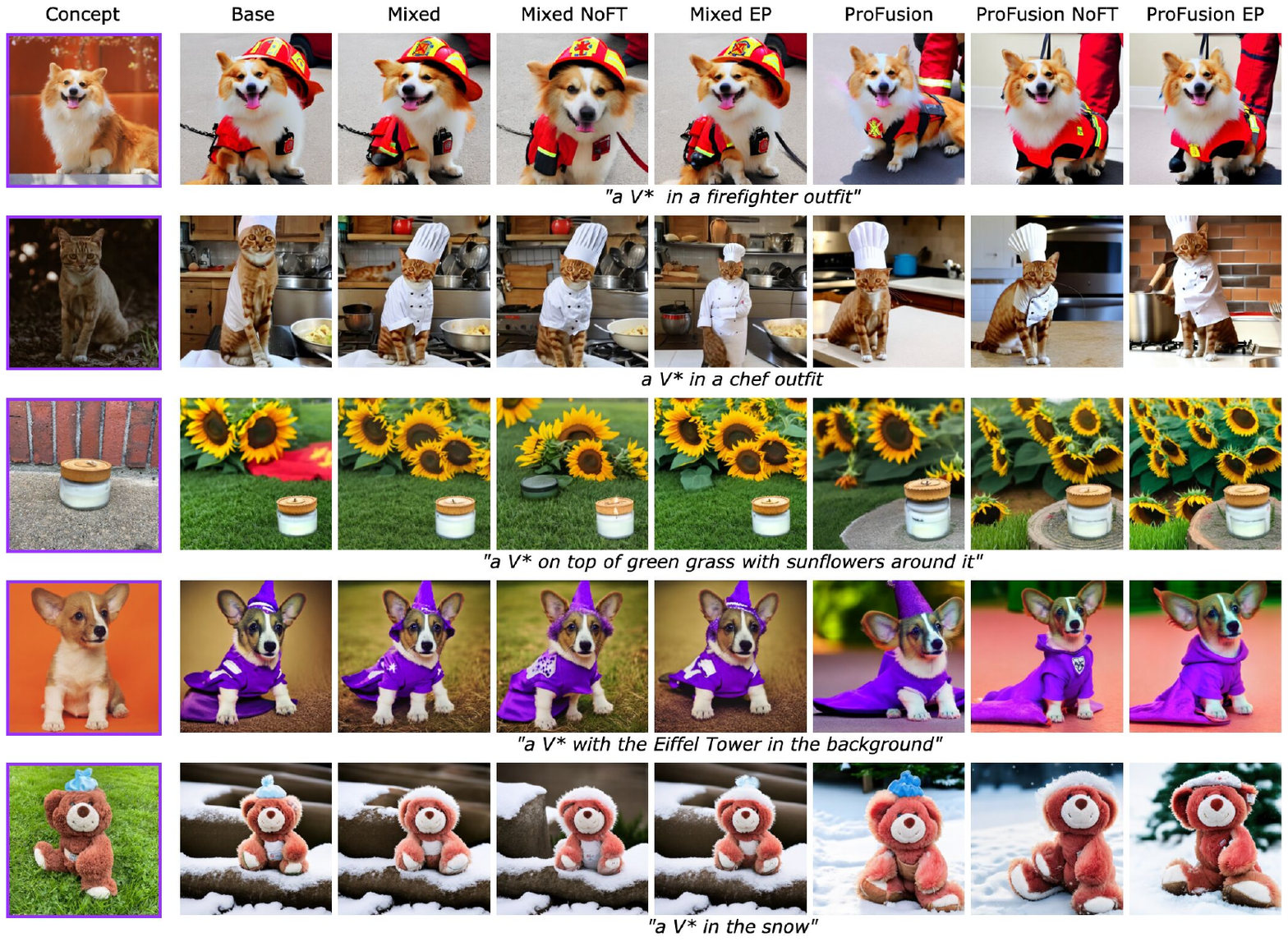}
    \caption{Examples of the generation outputs for Mixed and ProFusion sampling methods for their optimal metrics point in the Superclass, NoFT, and Empty Prompt (EP) setups.} \label{fig:examples_noft_ep}
\end{figure}

A key limitation of increased misalignment is the gradual erosion of superclass category information from generated images, which can lead to semantically inconsistent outputs. For instance, Figure~\ref{fig:examples_noft_ep} illustrates how the Mixed Empty Prompt setup, despite the strong animal prior in Base sampling, can produce human-like features in an image of a cat described as \textit{"in a chef outfit"}. This suggests that when superclass information is weakened, the model may introduce unexpected visual artifacts, impacting the fidelity of the intended concept.

Concept sampling (\ref{eq:concept_sampling}) can also be adjusted to better capture a concept’s visual characteristics, further decoupling fidelity from editability. For example, this can be achieved by (1) using the weights of a highly overfitted model (e.g., DreamBooth) or (2) selecting a prompt that omits contextual details, such as $p^{\hat{C}} = \textit{"a photo of V*"}$ instead of $p^{C} = \textit{"a V* with a city in the background"}$. Combining superclass sampling under NoFT or Empty Prompt with Base sampling configured via (1) or (2) could enhance both image and text similarity. We leave this direction for future work.


\clearpage

\begin{figure}[ht!]
  \centering
  \includegraphics[trim={0 5cm 0 5cm},clip,width=0.95\linewidth]{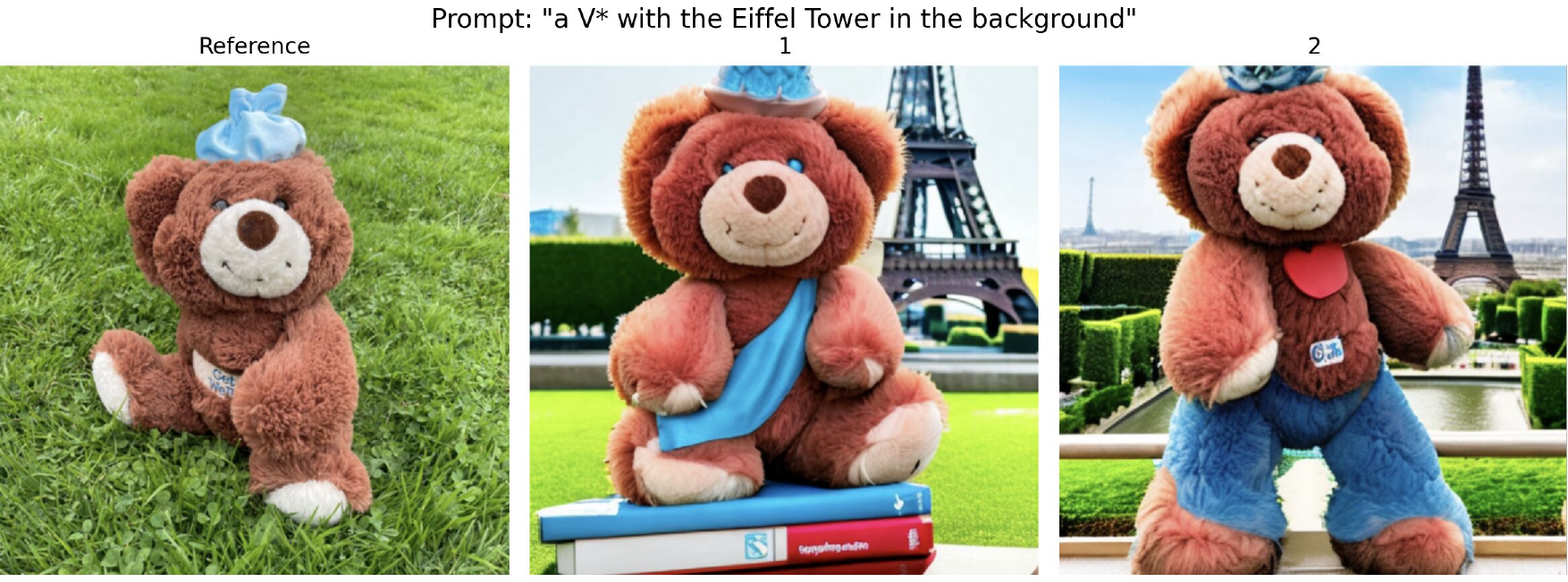}
  \caption{An example of a task in the user study}
  \label{fig:us_ex}
  \vspace{-0.19in}
\end{figure}

\section{Data preparation}\label{app:data}
For each concept, we used inpainting augmentations to create the training dataset. We took an original image and automatically segmented it using the Segment Anything model on top of the CLIP cross-attention maps. Then we crop the concept from the original image, apply affine transformations to it, and inpaint the background. We used $10$ augmentation prompts, different from the evaluation prompts, and sampled $3$ images per prompt, resulting in a total of $30$ training images per concept. We commit to open-source the augmented datasets for each concept after publication.

\section{User Study}\label{app:us}

An example task from the user study is shown in Figure~\ref{fig:us_ex}. In total, we collected 48,864 responses from 200 unique users for 16,000 unique pairs. For each task, users were asked three questions: 1) "Which image is more consistent with the text prompt?" 2) "Which image better represents the original image?" 3) "Which image is generally better in terms of alignment with the prompt and concept identity preservation?" For each question, users selected one of three responses: "1", "2", or "Can't decide."

\section{Complex Prompts Setting}\label{app:long_prompts}

We conduct a comparison of different sampling methods using a set of complex prompts. For this analysis, we collected 10 prompts, each featuring multiple scene changes simultaneously, including stylization, background, and outfit:

\adjustbox{max width=\linewidth}{
\begin{lstlisting}
live_long = [
  "V* in a chief outfit in a nostalgic kitchen filled with vintage furniture and scattered biscuit",
  "V* sitting on a windowsill in Tokyo at dusk, illuminated by neon city lights, using neon color palette",
  "a vintage-style illustration of a V* sitting on a cobblestone street in Paris during a rainy evening, showcasing muted tones and soft grays",
  "an anime drawing of a V* dressed in a superhero cape, soaring through the skies above a bustling city during a sunset",
  "a cartoonish illustration of a V* dressed as a ballerina performing on a stage in the spotlight",
  "oil painting of a V* in Seattle during a snowy full moon night",
  "a digital painting of a V* in a wizard's robe in a magical forest at midnight, accented with purples and sparkling silver tones",
  "a drawing of a V* wearing a space helmet, floating among stars in a cosmic landscape during a starry night",
  "a V* in a detective outfit in a foggy London street during a rainy evening, using muted grays and blues",
  "a V* wearing a pirate hat exploring a sandy beach at the sunset with a boat floating in the background",
]

object_long = [
  "a digital illustration of a V* on a windowsill in Tokyo at dusk, illuminated by neon city lights, using neon color palette",
  "a sketch of a V* on a sofa in a cozy living room, rendered in warm tones",
  "a watercolor painting of a V* on a wooden table in a sunny backyard, surrounded by flowers and butterflies",
  "a V* floating in a bathtub filled with bubbles and illuminated by the warm glow of evening sunlight filtering through a nearby window",
  "a charcoal sketch of a giant V* surrounded by floating clouds during a starry night, where the moonlight creates an ethereal glow",
  "oil painting of a V* in Seattle during a snowy full moon night",
  "a drawing of a V* floating among stars in a cosmic landscape during a starry night with a spacecraft in the background",
  "a V* on a sandy beach next to the sand castle at the sunset with a floaing boat in the background",
  "an anime drawing V* on top of a white rug in the forest with a small wooden house in the background",
  "a vintage-style illustration of a V* on a cobblestone street in Paris during a rainy evening, showcasing muted tones and soft grays",
]
\end{lstlisting}
}

\begin{figure}[h!]
  \centering
  \includegraphics[width=\linewidth]{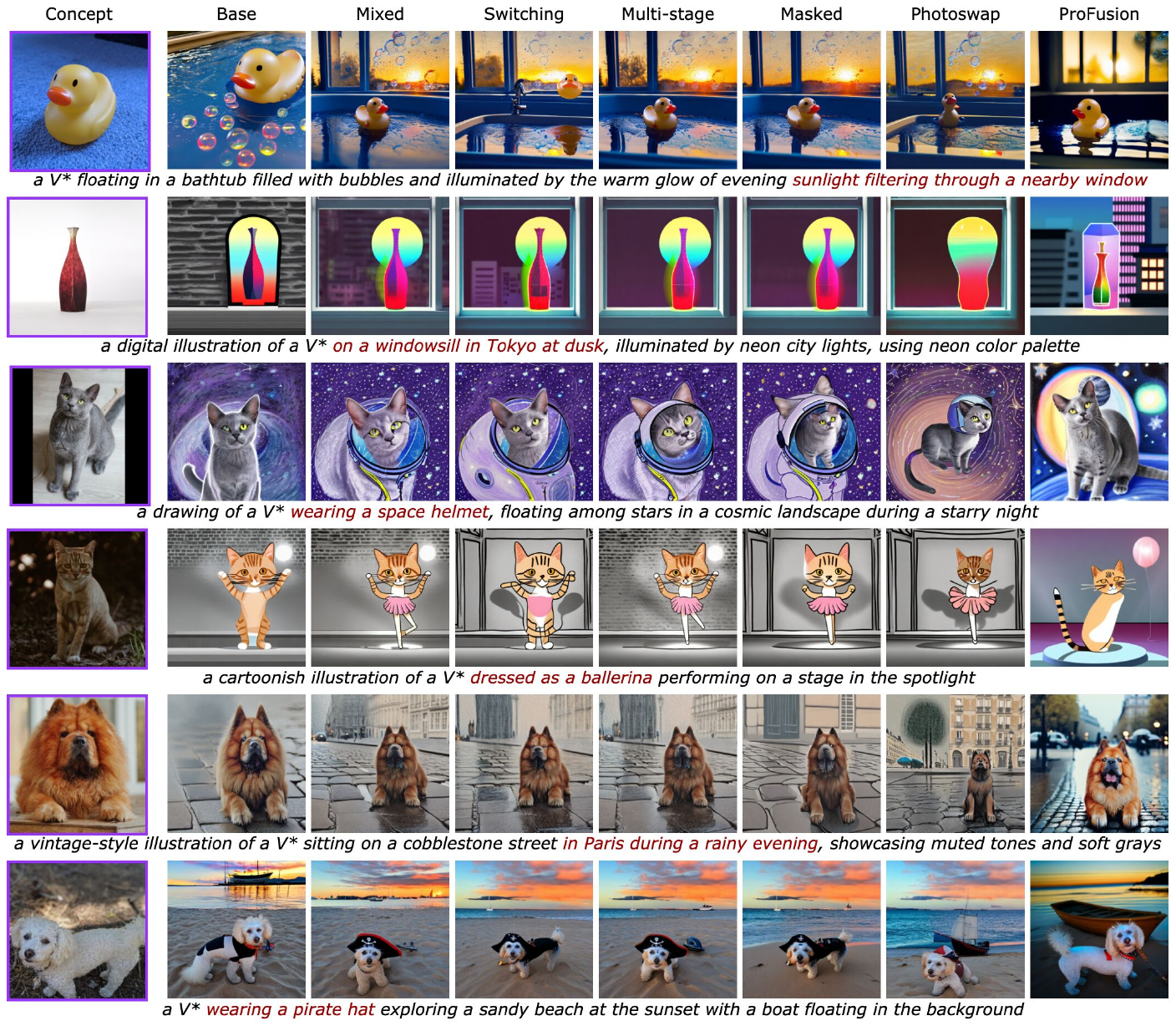}
  \caption{Additional examples of the generation outputs for different sampling methods with \textbf{complex prompts}. We highlight parts of the prompt that are missing in Base sampling while appearing in other methods.}
  \label{fig:add_long}
\end{figure}

The results of this comparison are presented in Figures~\ref{fig:add_long},~\ref{fig:add_long_metrics}. We observe that Base sampling may struggle to preserve all the features specified by the prompts, whereas advanced sampling techniques effectively restore them. The overall arrangement of methods in the metric space closely mirrors that observed in the setting with simple prompts.

\clearpage
\section{Dreambooth results}\label{app:dreambooth}

\begin{figure*}[!ht]
\centering
\begin{minipage}{.477\textwidth}
  \centering
  \includegraphics[trim={3cm 10cm 3cm 10cm},clip,width=\linewidth]{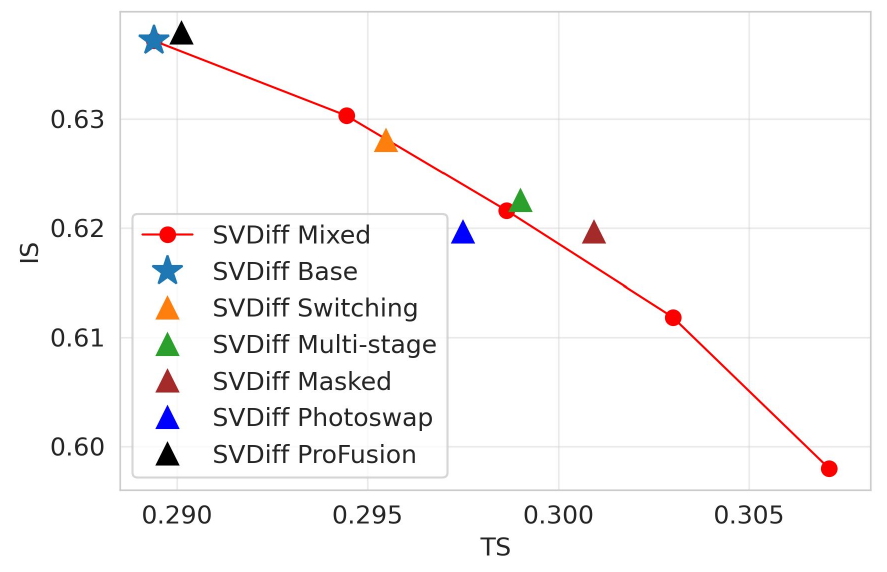}
  \caption{CLIP metrics for different sampling methods estimated on \textbf{complex prompts}.}
  \label{fig:add_long_metrics}
\end{minipage}
\hfill
\begin{minipage}{.477\textwidth}
  \centering
  \includegraphics[trim={3cm 10cm 3cm 10cm},clip,width=\linewidth]{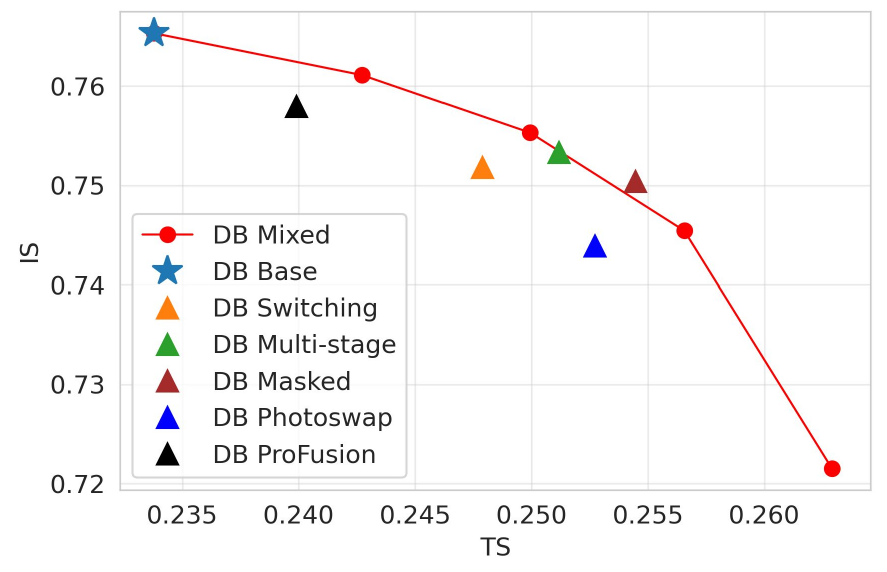}
  \caption{CLIP metrics for different sampling strategies on top of a Dreambooth fine-tuning method.}
  \label{fig:add_db_metrics}
\end{minipage}
\end{figure*}

\begin{figure}[h!]
  \centering
  \includegraphics[trim={0 1cm 0 1cm},clip,width=\linewidth]{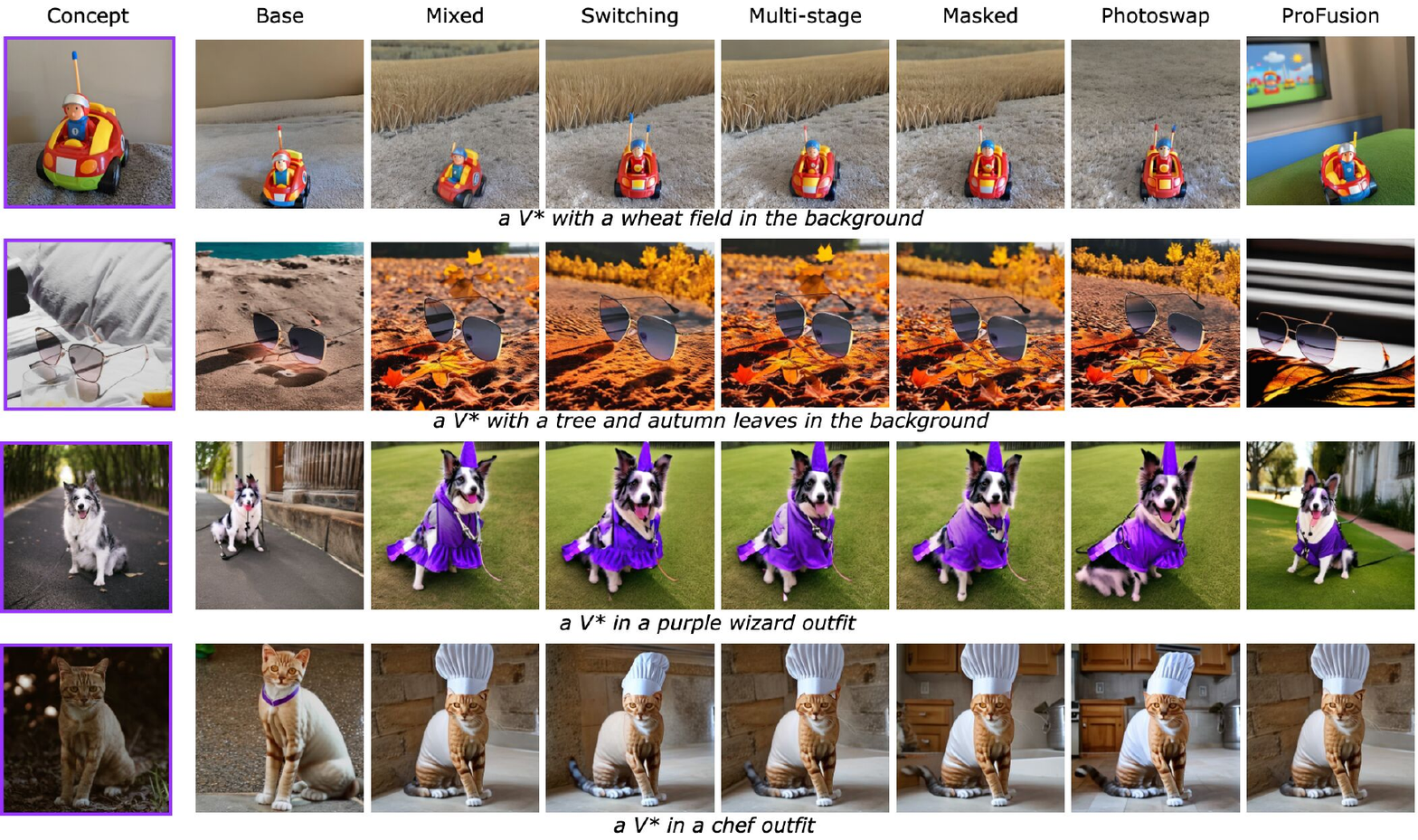}
  \caption{Additional examples of the generation outputs for different sampling methods on top of a Dreambooth fine-tuning method.}
  \label{fig:add_db}
\end{figure}

We conduct additional analysis of different sampling methods in combination with Dreambooth. Figure~\ref{fig:add_db_metrics} shows that Mixed Sampling still overperforms Switching and Photoswap,  while Multi-stage and Masked struggle to provide an additional improvement over the simple baseline. Figure~\ref{fig:add_db} shows that all methods allow for improvement TS with a negligent decrease in IS while Mixed Sampling provides the best IS among all samplings.

\section{PixArt-alpha \& SD-XL}\label{app:add_backbones}
We conducted a series of experiments using different backbones. For SD-XL~\cite{podell2023sdxlimprovinglatentdiffusion}, we used SVDDiff as the fine-tuning method, while PixArt-alpha~\citep{chen2023pixartalphafasttrainingdiffusion} employed standard Dreambooth training. Hyperparameters for Switching, Masked, and ProFusion were selected in the same manner as in the experiments with SD2.

\begin{figure}[h]
\centering
\begin{minipage}{.49\textwidth}
  \centering
  \includegraphics[trim={3cm 10cm 3cm 10cm},clip,width=\linewidth]{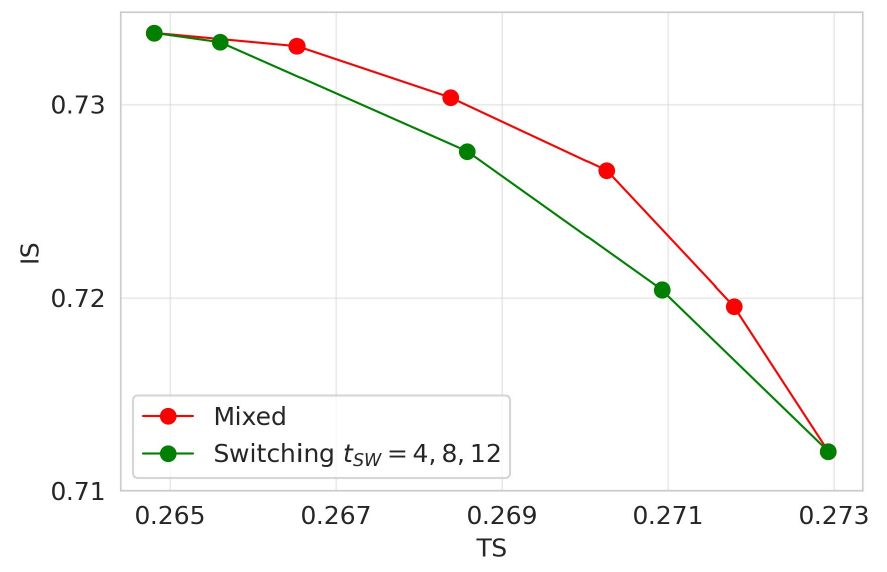}
  \captionof{figure}{CLIP metrics for different sampling methods estimated on PixArt model.}
  \label{fig:pixart}
\end{minipage}%
\hfill
\begin{minipage}{.49\textwidth}
  \centering
  \includegraphics[trim={3cm 10cm 3cm 10cm},clip,width=\linewidth]{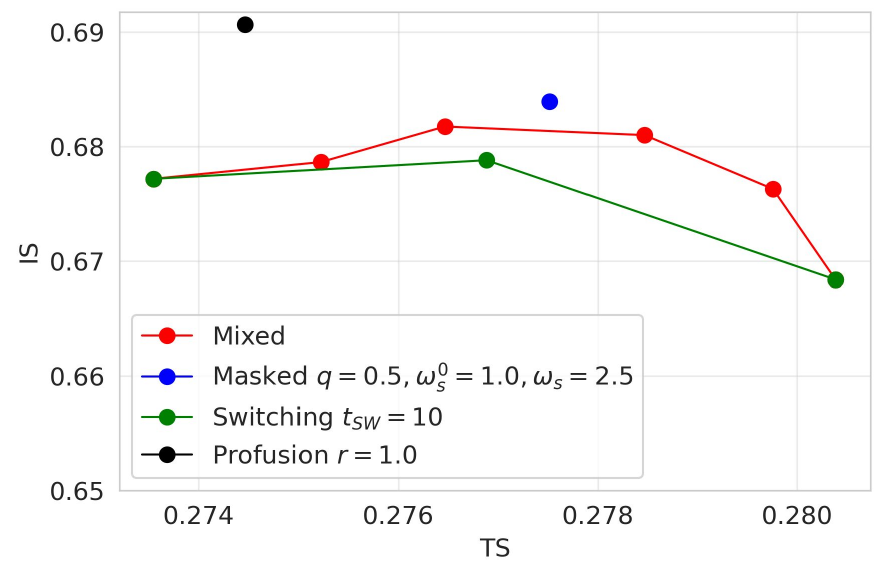}
  \captionof{figure}{CLIP metrics for different sampling methods estimated on SD-XL model.}
  \label{fig:sdxl}
\end{minipage}
\end{figure}

Figures~\ref{fig:pixart} and~\ref{fig:sdxl} demonstrate that Mixed Sampling follows a similar pattern to SD2, improving TS without a significant loss in IS. Notably, Mixed Sampling for SD-XL achieves simultaneous improvements in both IS and TS. ProFusion exhibits behavior consistent with SD2, enhancing IS more effectively than Mixed Sampling but performing worse at improving TS while also requiring twice the computational resources. 

\clearpage
\section{Cross-Attention Masks}\label{app:cross_attn}

\begin{figure}[h!]
  \centering
  \includegraphics[trim={3cm 0cm 3cm 0cm},clip,width=\linewidth]{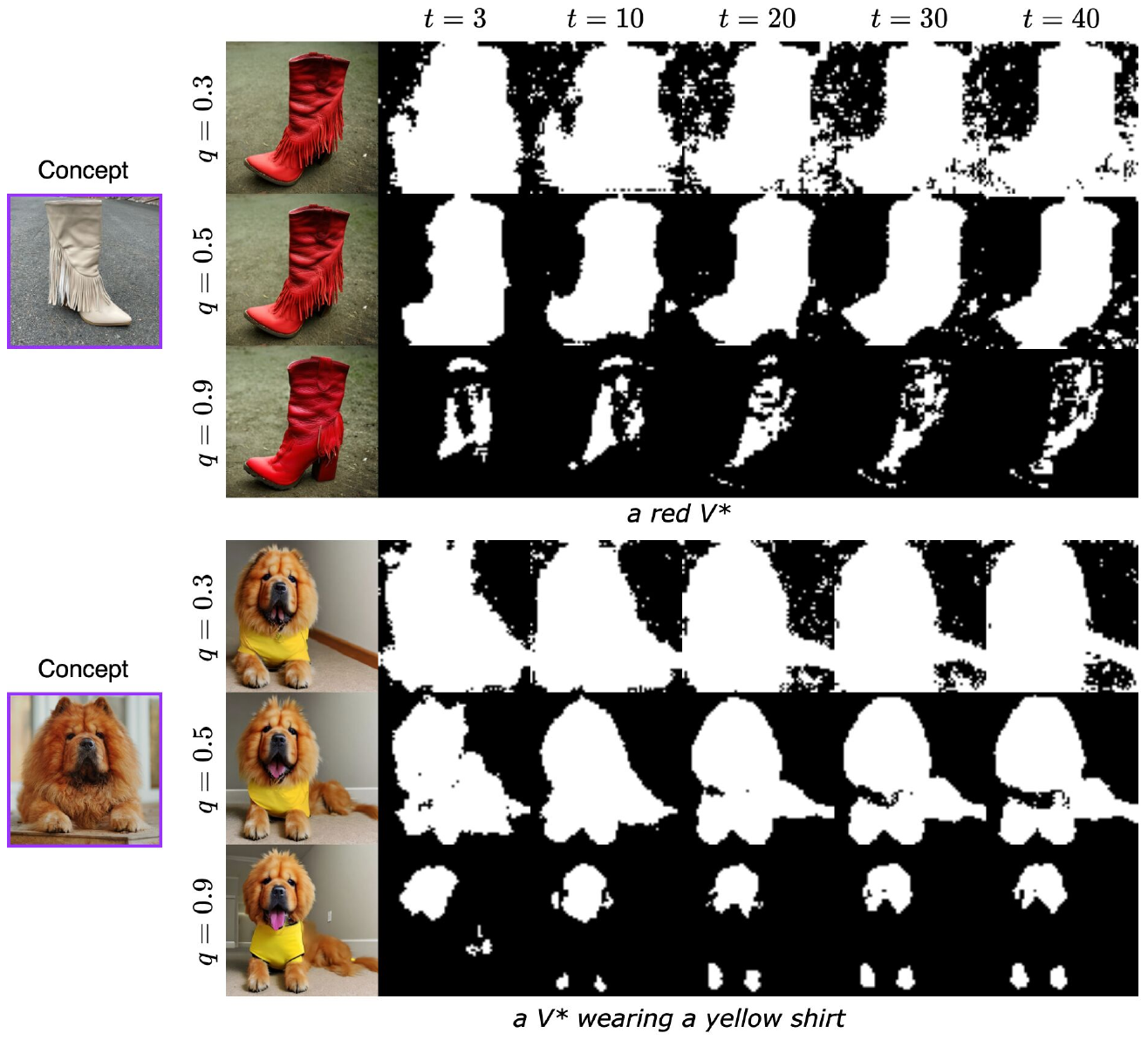}
  \caption{Visualization of the cross-attention masks for Masked sampling examples. Here, $q$ defines the thresholding quantile and $t$ the denoising step.}
  \label{fig:cross_attn_add_ex}
\end{figure}

\clearpage
\section{Additional Examples}\label{app:add_example}

\begin{figure}[h!]
  \centering
  \includegraphics[trim={0 2cm 0 2cm},clip,width=\linewidth]{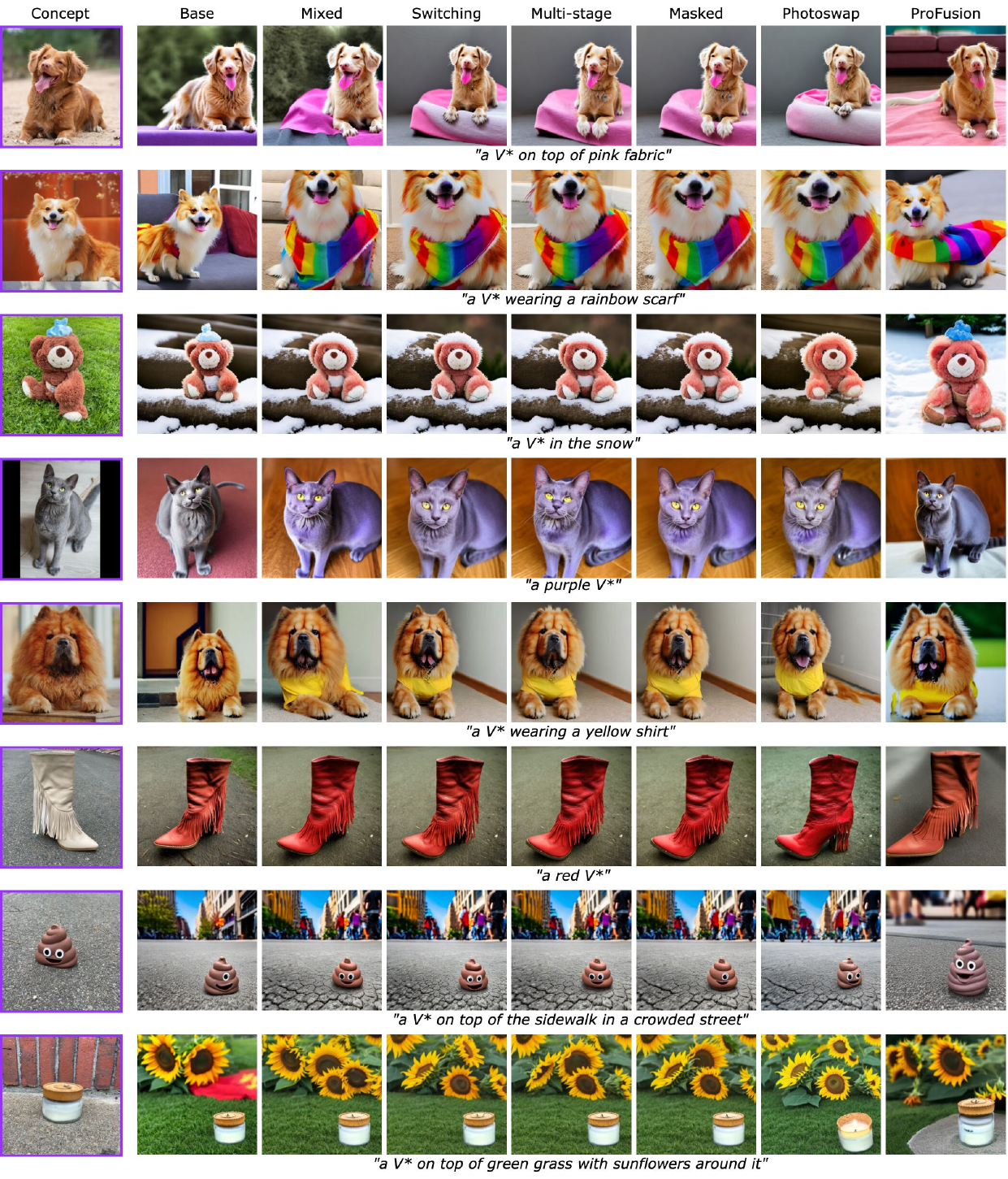}
  \caption{Additional examples of the generation outputs for different sampling methods.}
  \label{fig:add_ex}
\end{figure}

\begin{figure}[h!]
  \centering
  \includegraphics[trim={0 2cm 0 2cm},clip,width=\linewidth]{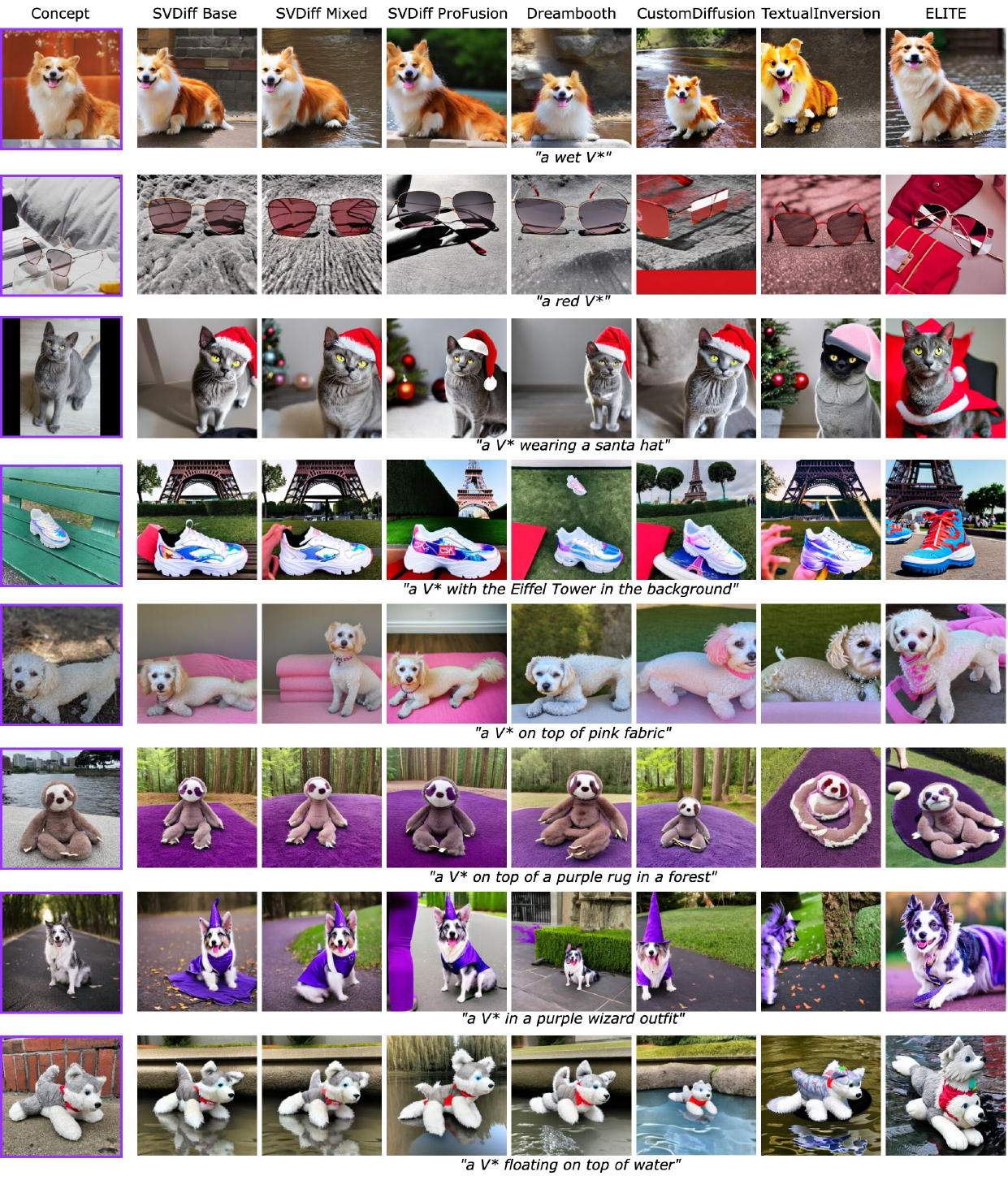}
  \caption{Additional examples of the generation outputs for Mixed and ProFusion sampling methods in comparison to the main personalized generation baselines.}
  \label{fig:add_ex_all}
\end{figure}

\clearpage
\section{DINO Image Similarity}\label{app:add_dino}

\begin{figure}[h]
\centering
\begin{minipage}{.49\textwidth}
  \centering
  \includegraphics[trim={3cm 10cm 3cm 10cm},clip,width=\linewidth]{imgs/profusion_photoswap.pdf}
\end{minipage}%
\hfill
\begin{minipage}{.49\textwidth}
  \centering
  \includegraphics[trim={3cm 10cm 3cm 10cm},clip,width=\linewidth]{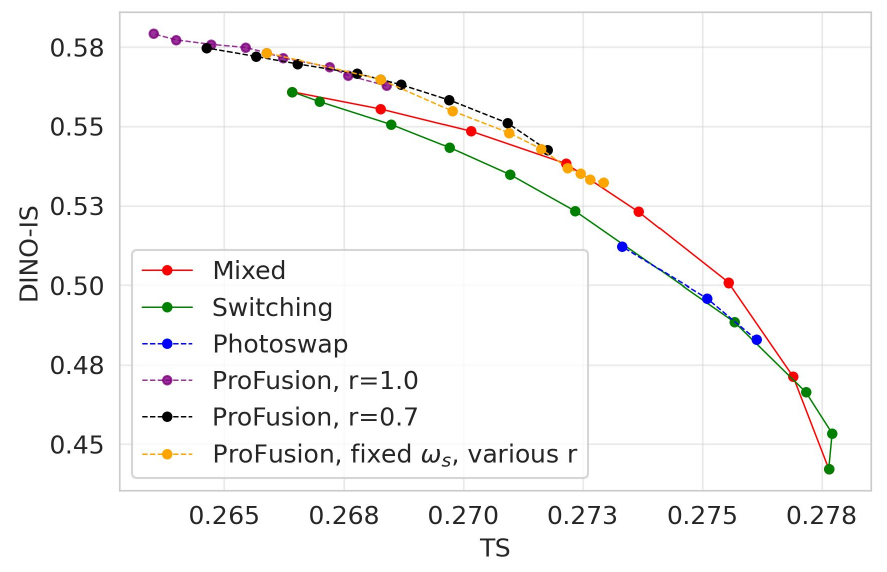}
\end{minipage}
\caption{Pareto frontiers curves for Photoswap~\citep{photoswap} and ProFusion~\citep{profusion}.}\label{fig:profusion_photoswap_dino}
\end{figure}

\begin{figure}[h]
\centering
\begin{minipage}{.49\textwidth}
  \centering
  \includegraphics[trim={3cm 10cm 3cm 10cm},clip,width=\linewidth]{imgs/all_methods.pdf}
\end{minipage}%
\hfill
\begin{minipage}{.49\textwidth}
  \centering
  \includegraphics[trim={3cm 10cm 3cm 10cm},clip,width=\linewidth]{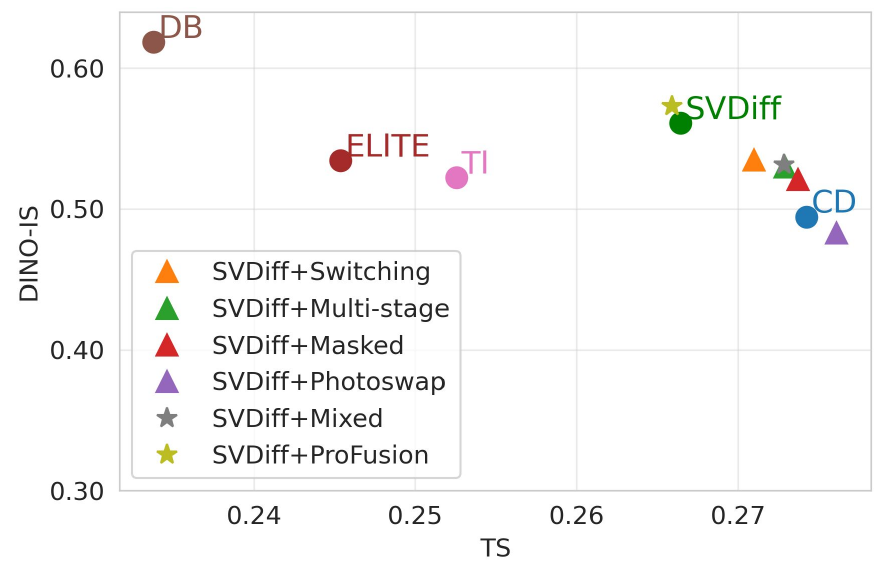}
\end{minipage}
\caption{The overall results of different sampling methods against main personalized generation baselines.}\label{fig:all_methods_dino}
\end{figure}

We compare CLIP-IS (left column) and DINO-IS~\citep{oquab2024dinov2learningrobustvisual} (right column) in Figures~\ref{fig:profusion_photoswap_dino},~\ref{fig:all_methods_dino}. We observe that despite the choice of metric, different sampling techniques and finetuning strategies have the same arrangement. The most noticeable difference is that SVDDiff superiority over ELITE and TI is more pronounced. That strengthens our motivation to select SVDDiff as the main backbone.

\end{document}